\documentclass[11pt]{article}

\usepackage[letterpaper,margin=1in,headheight=14pt]{geometry}
\usepackage[T1]{fontenc}
\usepackage{lmodern}
\usepackage{microtype}
\usepackage{fancyhdr}
\usepackage[round]{natbib}
\usepackage[hyphens]{url}
\usepackage{graphicx}
\usepackage{algorithm}
\usepackage{algorithmic}
\usepackage{booktabs}
\usepackage{amsmath}
\usepackage{amssymb}
\usepackage{multirow}
\usepackage{placeins}
\usepackage{xcolor}
\usepackage[hidelinks]{hyperref}

\usepackage{titling}
\predate{}\postdate{}
\posttitle{\par\end{center}\vspace{0.5em}}
\usepackage{abstract}
\usepackage{authblk}

\newcommand{\artFull}{Art-Based Reinforcement Training}

\usepackage{fontawesome5}
\usepackage{iftex}
\ifluatex
  \usepackage{emoji}
\else
  \newcommand{\emoji}[1]{\faPalette}
\fi

\newcommand{\todo}[1]{}

\begin{document}
\title{Fine-tuning Multi-modal LLMs with ART: \\
\emoji{artist-palette} \artFull}

\author[1]{Michal Chudoba}
\author[2]{Sergey Alyaev}
\author[1]{Petra Galuscakova}
\author[1]{Tomasz Wiktorski}
\affil[1]{University of Stavanger, Stavanger, Norway}
\affil[2]{NORCE Research, Bergen, Norway}
\affil[ ]{\texttt{\{michal.chudoba, petra.galuscakova, tomasz.wiktorski\}@uis.no}, \texttt{sergey.alyaev@norceresearch.no}}
\date{}
\maketitle

\begin{figure}[h!]
\centering
\setlength{\tabcolsep}{2pt}
\begin{tabular}{ccc}
\includegraphics[width=0.18\columnwidth]{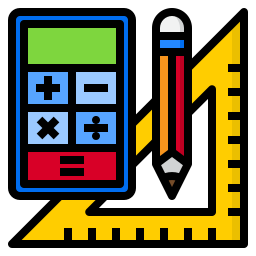} &
\includegraphics[width=0.18\columnwidth]{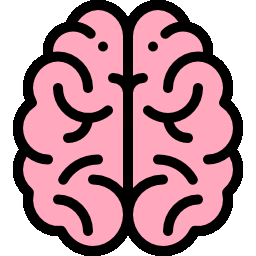} &
\includegraphics[width=0.18\columnwidth]{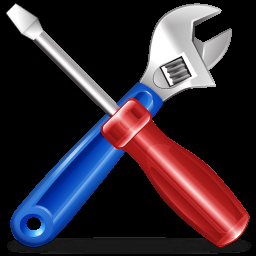} \\
Seed: math & Seed: brain & Seed: tool \\
\includegraphics[width=0.18\columnwidth]{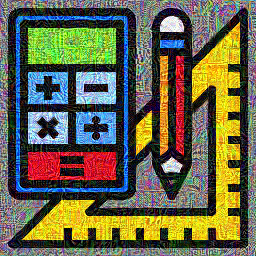} &
\includegraphics[width=0.18\columnwidth]{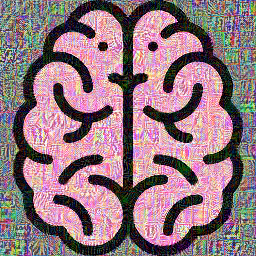} &
\includegraphics[width=0.18\columnwidth]{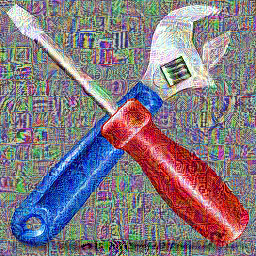} \\
ART: GSM8K & ART: DROP & ART: ToolMind \\
\textbf{+5.0 pp} &
\textbf{+16.9 pp} &
\textbf{+16.6 pp}\\
\end{tabular}
\caption{Top row: seed images (math~\citep{flaticon_math}, brain~\citep{flaticon_brain}, tool~\citep{freeiconspng_tool}). 
Bottom row: ART images for Qwen3.5-0.8B on GSM8K, DROP, and ToolMind (seed 1) and their percentage-point gain over text-only baselines.}
\label{fig:optimized_images}
\end{figure}

\begin{abstract}
Adapting Large Language Models (LLMs) to specific tasks usually involves fine-tuning: Low-Rank Adaptation (LoRA) edits the weights, while soft prompting prepends trainable virtual tokens. Both change how an already-optimized model is served, so serving many task-specific adapters at high throughput requires adapter-aware infrastructure.
We propose Art-Based Reinforcement Training (\textbf{ART}), which optimizes a single auxiliary input image to specialize a frozen Multimodal Large Language Model (MLLM) for a specific task.
At inference time, ART simply includes this image in each standard multimodal prompt, requiring no changes to the serving engine.
We compare ART with LoRA and soft prompting across five reasoning benchmarks for four small MLLMs from three architectures. 
While different methods perform best for different tasks, ART ranks first or second in 10 of 20 benchmark-MLLM pairs, with its best results on the two math word problem benchmarks (GSM8k and SVAMP). 
It is significantly ahead of the next best adapter in 2 cases, within the noise of the next best adapter in 6, and significantly above the text-only baseline in 18. 
ART therefore reaches accuracy close to LoRA and soft prompts while serving faster than LoRA. 
In the vLLM serving stack, ART reduces throughput by 6\% to 14\% relative to text-only requests, compared to LoRA's 13\% to 93\% slowdown.
Beyond efficient serving, ART artifacts are plain images, making them easy to store, distribute, and share. This approach keep task-specific adaptation  outside the model weights, so the same frozen model can support different tasks without separate model variants.

Code for verifying the method is available at \url{https://github.com/jinymusim/ART}.
\end{abstract}

\section{Introduction}
Modern MLLMs process text, image, and video input natively \citep{liu2023visual,liu2023improved,qwen35blog}, but many tasks for which they are deployed, such as mathematical reasoning, code execution, question answering, and tool use, are communicated by text-only instructions. The visual channel stays unused.
Task-specific fine-tuning yields substantial gains on these tasks~\citep{bucher2024fine}, especially for smaller models.
LoRA \citep{hu2021lora} has become the dominant Parameter-Efficient Fine-Tuning (PEFT) method in practice, and soft prompting \citep{lester2021power} offers an input-space alternative that supplies the fine-tuning information as virtual tokens alongside the prompt.
Both, however, add work to an already-optimized engine such as vLLM \citep{kwon2023efficient}. Serving many task-specific adapters at high throughput requires dynamic weight loading and per-adapter kernels for LoRA. Furthermore, the client must compute a precomputed-embedding input for the soft prompts.

We introduce \textbf{ART} (\artFull), which \textbf{repurposes the visual input channel} of MLLMs as a \textbf{non-invasive interface for task adaptation}. 
More generally, ART explores whether the input itself can be used to adapt a frozen model, rather than changing the model weights.
Rather than modifying model weights or inserting virtual tokens, ART \textbf{learns a task-specific input image} that is routed through the standard vision pathway.
During training, the pixel array is the only trainable parameter and gradients flow through the frozen model. 
For our training, we use a varient of reinforcement-learning policy optimization method, but the objective is replaceable.
During inference, the model stays frozen and the image is attached to each request. The serving engine treats it as a standard multimodal request, with no custom weight managers or specialized kernels.
Figure~\ref{fig:optimized_images} shows examples. The optimized images resemble their seeds with a high-frequency, task-specific structure added.

\paragraph{Contributions.}
\begin{itemize}
  \item We introduce \textbf{ART}, a method that adapts a frozen MLLM by optimizing only a task-specific input image, deployed as a standard RGB PNG file that can be visually styled for the task.
  \item We compare ART with LoRA and soft prompts under a unified policy optimization loop on twenty cases: five reasoning benchmarks for four small MLLMs (0.8--2B). ART reaches accuracy close to the other two adaptations.
  \item We benchmark multi-tenant serving speed. ART runs faster than LoRA and, unlike soft prompts, needs no infrastructure beyond the standard multimodal request.
\end{itemize}

\section{Related Work}
\label{sec:related_work}

\paragraph{Visual Prompt Tuning, Reprogramming, and Adversarial Steering.}
Tuning the input pixels of a vision model to perform new tasks was popularized by Adversarial Reprogramming \citep{elsayed2018adversarial}, which showed that adding a single optimized perturbation to ImageNet classifiers could repurpose them for out-of-domain tasks such as MNIST digit classification.
Visual prompting \citep{bahng2022exploring,wu2022unleashing} extended this by learning visual prompts in the pixel space of frozen CLIP vision encoders.
Visual Prompt Tuning (VPT) \citep{jia2022visual} introduced continuous learnable tokens in the intermediate layers of a Vision Transformer.
Transferable Visual Prompting (TVP) \citep{zhang2024exploring} learns a pixel-space perturbation of the input images that improves frozen MLLMs on vision tasks such as object recognition and counting, and transfers across models.
Recent investigations into MLLM security have revealed that the visual channel exerts an outsized influence on text generation.
\citet{qi2024visual} showed that a single adversarial image can jailbreak aligned models.
Bailey et al. introduced Image Hijacks \citep{bailey2024image}, which use behavioral matching to train images that force specific textual outputs.
VISOR \citep{phute2025visor,balakrishnan2025visorpp} trains a universal image that reproduces a steering vector in the model's activations to control behavior such as refusal.
Unlike the reviewed methods, ART uses the visual channel constructively to raise task accuracy. It trains a standalone image end to end on the task reward and treats the pixels as adapter weights rather than as a bounded perturbation or an activation target.

\paragraph{Continuous Soft Prompting.}
Soft Prompting and Prefix-Tuning \citep{lester2021power,li2021prefix} optimize a sequence of virtual embedding vectors prepended to the frozen LLM input.
Discrete prompt optimizers such as RLPrompt \citep{deng2022rlprompt}, OPRO \citep{yang2024large}, PEZ \citep{wen2023hard}, and GEPA \citep{agrawal2026gepa} search directly in plain text.
\citet{patel2025interpretable} propose faithfulness and scrutability criteria and training objectives that trade a little accuracy for more interpretable soft prompts.
Unlike these, ART is an input-space prefix that passes through the standard vision encoder. Its artifact is a plain PNG image.

\paragraph{Parameter-Efficient Fine-Tuning and LoRA.}
PEFT methods adapt a pretrained model by training a small number of parameters while the base weights stay frozen.
LoRA \citep{hu2021lora} is the dominant PEFT method. It adds a trainable low-rank update $BA$ to selected weight matrices, so a rank-$r$ adapter for a $d \times k$ matrix trains only $r(d+k)$ parameters.
LoRA is supported by the Transformer Reinforcement Learning (TRL) library \citep{vonwerra2022trl} and by vLLM.
Unless the adapter is merged into the weights, every decoding step runs the extra low-rank product, and a batch that mixes tenants needs a different adapter per request.
Systems such as S-LoRA \citep{sheng2024slora} and Punica \citep{chen2024punica} serve many adapters at once by batching them through custom kernels.
These systems reduce the cost but do not remove it, because they still need adapter-aware weight management and LoRA kernels on top of the base engine.

\paragraph{Reinforcement Learning for LLM Reasoning.}
Group Relative Policy Optimization (GRPO) was introduced by DeepSeekMath \citep{shao2024deepseekmath} and popularized by DeepSeek-R1 \citep{deepseek2025deepseek} to improve reasoning in LLMs with rewards that can be checked automatically, such as a correct final answer.
Proximal Policy Optimization (PPO) \citep{schulman2017proximal} estimates the advantage of each completion with a learned critic, which is a second model of similar size to the policy.
GRPO removes the critic. It samples a group of $G$ completions for each prompt, scores each completion, and uses its reward minus the group mean, divided by the group standard deviation, as the advantage.
The policy is then updated with a clipped PPO-style objective, usually with a KL penalty to a reference model.
Dr.~GRPO (GRPO Done Right) \citep{liu2025drgrpo} shows that both normalizations in GRPO bias the optimization. Dividing by the group standard deviation over-weights prompts that are almost always solved or almost never solved, and averaging the loss per sequence favors short correct answers and long incorrect ones. It removes the standard-deviation term and normalizes the loss by a constant number of tokens instead.
We train all three methods with Dr.~GRPO.

\section{ART Framework}
\label{sec:art}

In this section, we formulate the ART fine-tuning framework.
Rather than adjusting model weights $\theta$, ART freezes the MLLM $M_{\theta}$ and optimizes the input image $X_{\text{pixel}} \in \mathbb{R}^{3 \times H \times W}$.
The following two subsections specify the parametrization and optimization details.
Section~\ref{sec:properties} summarizes the practical properties of the resulting artifacts.

\subsection{Pixel-Space Parameterization}
\label{sec:parameterization}

An MLLM processes images through a Vision Transformer (ViT) encoder.
The raw 8-bit image is resized, normalized, and split into visual patches, which are then projected into the shared embedding space alongside text tokens.
The critical property we exploit is that this pipeline is continuous and differentiable with respect to the input pixels.

We parameterize the learnable image directly in pixel space.
Given a seed image $X_{\text{pixel}}^{(0)}$ (an 8-bit RGB image), we initialize the trainable tensor $X \in \mathbb{R}^{1 \times 3 \times H \times W}$ by rescaling to the unit interval:
\begin{equation}
X = X_{\text{pixel}}^{(0)} \;/\; 255 \;\in\; [0, 1]^{1 \times 3 \times H \times W}.
\end{equation}
If no seed is provided, every value of $X$ is drawn from a normal distribution with mean $0.5$ and standard deviation $0.05$ and clamped to $[0, 1]$.
After every optimizer step we re-clamp $X$ back into $[0, 1]$, which keeps the rendered image in the valid range while leaving the optimizer free to take unconstrained steps.

At any point during training the 8-bit image is recovered by quantizing $X$:
\begin{equation}
X_{\text{pixel}} = \text{round}\big(X \cdot 255\big).
\end{equation}
For the backward pass we feed the appropriately normalized tensor directly into the frozen model.
The normalization statistics follow each model family's processor and may be ImageNet \citep{deng2009imagenet}, OpenAI CLIP, or standard ($\mu=\sigma=0.5$) statistics, depending on the backbone. We use whichever statistics the shipped processor declares in its configuration. To keep the backward pass consistent with the 8-bit rollout image, the forward value is rounded to 8-bit precision with the straight-through estimator $\tilde{X} = X + \mathrm{sg}\big(\text{round}(255X)/255 - X\big)$, where $\mathrm{sg}$ stops the gradient. The backward pass therefore sees the same image as vLLM, while gradients flow through the continuous pixel tensor.
Gradient updates are computed with respect to $X$ using Lion \citep{chen2024lion} ($\beta_1 = \beta_2 = 0.95$, weight decay $0$) in full $32$-bit precision to ensure numerical stability during backpropagation.

\begin{algorithm}[t]
\caption{Art-based Reinforcement Training (ART).}
\label{alg:art}
\begin{algorithmic}[1]
\REQUIRE Frozen MLLM $M_\theta$
\REQUIRE Training dataset $D$
\REQUIRE Differentiable loss function $\mathcal{L}$
\REQUIRE Optimizer $\mathcal{O}$
\STATE Allocate $X \in \mathbb{R}^{1 \times 3 \times H \times W}$
\IF{seed image $X_{\text{pixel}}^{(0)}$ provided}
  \STATE $X \leftarrow X_{\text{pixel}}^{(0)} / 255$ \COMMENT{rescale seed to $[0,1]$}
\ELSE
  \STATE $X \leftarrow \mathrm{clamp}(0.5 + 0.05\,\epsilon,\, 0,\, 1)$ with $\epsilon \sim \mathcal{N}(0, 1)$ per value
\ENDIF
\FOR{step $= 1, 2, \dots$}
  \STATE \COMMENT{Pass A. Rollout (via serving engine)}
  \STATE Quantize to 8-bit: $X_{\text{pixel}} \leftarrow \text{round}(X \cdot 255)$
  \STATE Produce $G$ outputs $\{o_{i,j}\}_{j=1}^G$ from $M_\theta(Q_{i}, X_{\text{pixel}})$ for each query $Q_{i}$
  \STATE Score outputs with rewards $\{r_{i,j}\}$ and compute advantages $\{A_{i,j}\}$ (or likelihoods for SFT)
  \STATE \COMMENT{Pass B. Backward (via frozen model)}
  \STATE Substitute $\tilde{X}$ as the visual input
  \STATE Compute the log-probabilities $\log \pi_X(o_{i,j,t} \mid Q_i, o_{i,j,<t})$ of the sampled tokens in one forward pass
  \STATE Compute loss $\mathcal{L}$ (Eq.~\ref{eq:loss} for Dr.~GRPO)
  \STATE $g \leftarrow \nabla_{X} \mathcal{L}$ (all model weights $\theta$ frozen)
  \STATE $X \leftarrow \mathcal{O}(X, g)$
  \STATE $\mathrm{clamp}(X,\, 0,\, 1)$ in place
\ENDFOR
\RETURN 8-bit PNG image $X_{\text{pixel}} = \text{round}(X \cdot 255)$
\end{algorithmic}
\end{algorithm}

\subsection{Reward-Driven Optimization in Pixel Space}
\label{sec:optimization}

During optimization, we wish to maximize the expected task accuracy of the outputs generated by $M_\theta(Q, X_{\text{pixel}})$, where $Q$ is a text query and $X_{\text{pixel}}$ is the prepended 8-bit image.
Because the optimization target $X$ is a plain tensor, the objective is agnostic to the training algorithm.
Algorithm~\ref{alg:art} shows the full pseudo-code of the ART training.
Our reference implementation uses Dr.~GRPO~\citep{liu2025drgrpo} training, and its two-pass loop is described below.

\paragraph{Pass A. Rollout and Advantage Estimation.} For a batch of prompts $\{Q_1, \dots, Q_B\}$, we quantize the continuous pixel array $X$ to an 8-bit RGB image $X_{\text{pixel}}$ and pass it to a vLLM engine, which samples $G$ completions per query ($o_{i,j} \sim M_\theta(Q_i, X_{\text{pixel}})$).
Each completion is scored by the dataset-specific reward $R(o_{i,j}, y_i)$ against the ground truth $y_i$.
We compute group-relative advantages following Dr.~GRPO~\citep{liu2025drgrpo}: mean-centered rewards without the standard-deviation denominator, which avoids amplifying noise from near-zero-variance groups early in training.
Completions that hit the length limit without emitting an end-of-sequence token are masked out.

\paragraph{Pass B. Backward Step.} We substitute the normalized tensor $(\tilde{X} - \mu)/\sigma$ into a frozen copy of the model and backpropagate the token-level-normalized policy gradient into $X$:
\begin{equation}
\label{eq:loss}
\begin{split}
\mathcal{L}(X) = - \frac{1}{T_{\max} N_{\text{comp}}} \sum_{i,j} \sum_{t=1}^{|o_{i,j}|} \\
\log \pi_{X}\!\left(o_{i,j,t} \mid Q_i, o_{i,j,<t}\right) \, A_{i,j},
\end{split}
\end{equation}
where $\pi_{X}$ is the frozen model's per-token probability under the current image, $A_{i,j}$ is the group-relative advantage of completion $o_{i,j}$ computed in Pass~A, $T_{\max}$ is the maximum completion length, and $N_{\text{comp}}$ is the number of completions in the step.
The token-level normalization follows Dr.~GRPO and removes the response-length bias of per-sequence normalization.
Because each rollout is consumed by a single gradient step, the sampling and differentiated policies match up to min-p truncation and numerical differences between vLLM and base model. The objective therefore reduces to a mean-centered policy gradient with no importance-sampling correction, KL penalty, or reference model.

Appendix~\ref{sec:methods} shows that supervised fine-tuning (SFT) \citep{vonwerra2022trl} can also produce ART images.

\subsection{Properties of the ART Artifacts}
\label{sec:properties}

The defining property of ART is that the learned artifacts are images (Figure~\ref{fig:optimized_images}), so they are native inputs to a pre-optimized MLLM.
Unlike LoRA, which couples the adapter to a specific weight decomposition, or soft prompting, which injects virtual embedded vectors, the ART artifacts live entirely in pixel space.
This native-input representation has two practical consequences.

First, the artifact is naturally deployable and compressible.
The deployed image is a standard 8-bit RGB PNG (3 bytes per pixel), analogous to INT8 model-weight quantization.
In Section~\ref{sec:resultsStorage} we investigate how the information accumulates in the image, while Section~\ref{sec:robustness} checks if it can survive lossy image compression.

Second, because the artifact is processed through the standard vision pathway, the adapted model remains completely frozen, and serving treats ART-conditioned inference as a standard multimodal request.

\section{Experimental Setup}
\label{sec:experiments}

To isolate the unique behavior of ART visual prompting we run a dense grid of baselines and fine-tuning methods on a single NVIDIA A100-40GB GPU.

\subsection{Benchmark Datasets}
\label{sec:datasets}

We evaluate ART across five open-ended, free-form reasoning benchmarks that span distinct skills. These are math word problems (GSM8K \citep{gsm8k}, SVAMP \citep{patel2021svamp}), discrete reasoning over text (DROP \citep{dua2019drop}), synthetic logical inference (bAbI \citep{weston2015babi}), and structured tool use (ToolMind \citep{nanbeige2025toolmind}).
All reported evaluation numbers use strict binary exact-match scoring. During training, every benchmark uses a lightly shaped reward. Appendix~\ref{sec:datasets_detail} gives the scorer and the shaped reward of each benchmark.

\subsection{MLLMs}
We test three backbone architectures at local scale.
Qwen3.5-0.8B is a small text-plus-vision MLLM from the Qwen family \citep{qwen35blog}.
Qwen3.5-2B complements it as a larger member of the same family.
Qwen2-VL-2B-Instruct~\citep{Qwen2VL} and LFM2-VL-1.6B~\citep{liquidai2025lfm2} are two further open-weight multimodal architectures with distinct vision towers and patchification schemes, which lets us probe whether ART generalizes beyond a single backbone.
We also train ART and the soft prompt on three larger models, Qwen2-VL-7B-Instruct, GLM-4.6V-Flash, and Gemma-3-12B-it (Appendix~\ref{sec:large_models}).
All visual inputs are 3-channel RGB at $256 \times 256$ unless otherwise noted, which maps to 64--81 visual tokens after patchification and spatial merge, depending on the backbone (per-model token counts in Appendix~\ref{sec:imgsize}).

\subsection{Fine-Tuned Setups}
\label{sec:fineTuningSetup}
\begin{description}
  \item[LoRA] Language-model LoRA fine-tuning with TRL \citep{vonwerra2022trl}.
  \item[Soft Prompt] A learnable embedding prefix of the same token count as the image produces after patchification (64--81, model-dependent), initialized by sampling rows from the model's input-embedding table.
  \item[Optimized Image (ART)] Our proposed method. The learnable pixel parameter is initialized from a fixed seed image and trained via Algorithm~\ref{alg:art}. We assign semantically meaningful initial imagery as seeds.
\end{description}

For the main results, all three fine-tuning methods share the same Dr.~GRPO objective, group size, temperature, sampling parameters, and reward functions. 
However, Appendix~\ref{sec:methods} extends the ART testing to another training objective.

The optimizer and learning rate are method-appropriate. ART uses Lion for the image, while the soft prompt and LoRA use AdamW (Appendix~\ref{sec:counterparts} gives their implementation details). Appendix~\ref{sec:config} lists the full fine-tuning configurations.
The main difference between the methods, however, is where the task signal is stored: in pixel space for ART (through the vision encoder), in embedding space for the soft prompt, and in weight space for LoRA (rank-8 language-model adapters). 
This location also determines how each artifact is served.

\paragraph{Data and checkpoint selection.}
Each method trains for one epoch on the first 700 training examples of a benchmark, of which the last 192 form a validation probe.
Every 20 steps we score the probe with the strict metric and keep the latest checkpoint within a standard error of the best probe accuracy. 
Examples of full training curves 
are available in Appendix~\ref{sec:training_dynamics}.
We train ART with three seeds on every model, and all three methods with three seeds on Qwen3.5-0.8B. The seed changes the prompt order, the sampled rollouts, and the random initialization of the soft prompt and LoRA. 
Analyses of a single image (transfer, robustness, and pixel statistics) use the seed-1 image.

\subsection{Non-Fine-Tuned Controls}
\begin{description}
  \item[Baseline] Standard text-only generation. No image prefix is provided.
  \item[Random Image] Prepend a freshly sampled random $256 \times 256$ RGB image, unique to each query. This controls for the mere presence of visual tokens.
  \item[Fixed Seed Image] The static, unoptimized image prepended to every query (also used as the coresponding training seed). 
  \item[Random String] Prepend random text tokens to every query, matched in count to the number of visual tokens. This serves as a text-space token-overhead baseline for the random-image condition.
\end{description}
These controls separate the effect of ``extra tokens'' from the effect of optimized 

\section{Results and Discussion}
\label{sec:results}

\subsection{ART fine-tuning performance}
\label{sec:main_results}

Table~\ref{tab:main_results} reports the main comparative results of ART's performance.
It gives the accuracy of the baseline, the controls, and the three fine-tuning methods on all five benchmarks, with 95\% bootstrap confidence intervals.
ART on every model and all methods on Qwen3.5-0.8B are averaged over three training seeds.

\begin{table*}[!t]
\centering
\caption{Accuracy (\%) with 95\% bootstrap CI half-widths. ART on every model and all methods on Qwen3.5-0.8B are three-seed means, with the seed standard deviation in parentheses and a CI over seeds and examples. Best in \textbf{bold}, second best \underline{underlined}.}
\label{tab:main_results}
\small
\setlength{\tabcolsep}{3pt}
\resizebox{\textwidth}{!}{\begin{tabular}{llccccc}
\toprule
\textbf{Model} & \textbf{Condition} & \textbf{bAbI} & \textbf{DROP} & \textbf{GSM8K} & \textbf{SVAMP} & \textbf{ToolMind} \\
\midrule
\multirow{7}{*}{LFM2-VL-1.6B}
& Baseline & 3.2 $\pm$ 0.8 & 21.1 $\pm$ 1.8 & 21.0 $\pm$ 2.2 & 34.0 $\pm$ 5.3 & 6.5 $\pm$ 1.1 \\
& Random Image & 4.6 $\pm$ 0.9 & 22.2 $\pm$ 1.8 & 24.8 $\pm$ 2.3 & 45.0 $\pm$ 5.7 & 5.6 $\pm$ 1.0 \\
& Fixed Seed Image & 5.9 $\pm$ 1.0 & 23.4 $\pm$ 1.9 & 18.6 $\pm$ 2.0 & 21.3 $\pm$ 4.5 & 6.2 $\pm$ 1.0 \\
& Random String & 2.9 $\pm$ 0.7 & 21.7 $\pm$ 1.8 & 32.4 $\pm$ 2.6 & 48.3 $\pm$ 5.7 & 6.1 $\pm$ 1.0 \\
\cmidrule{2-7}
& Soft Prompt & \textbf{52.6} $\pm$ 2.2 & \textbf{52.9} $\pm$ 2.2 & \textbf{64.7} $\pm$ 2.6 & \textbf{77.3} $\pm$ 4.7 & \textbf{94.2} $\pm$ 1.0 \\
& Optimized Image (ART) & \underline{40.8} $\pm$ 2.4\,{\scriptsize(1.3)} & \underline{40.8} $\pm$ 2.4\,{\scriptsize(1.3)} & 60.0 $\pm$ 2.5\,{\scriptsize(1.0)} & 73.0 $\pm$ 4.9\,{\scriptsize(2.1)} & 8.7 $\pm$ 1.6\,{\scriptsize(1.3)} \\
& LoRA & 33.1 $\pm$ 2.1 & 28.8 $\pm$ 2.0 & \underline{60.5} $\pm$ 2.7 & \underline{74.0} $\pm$ 4.8 & \underline{15.7} $\pm$ 1.6 \\
\midrule
\multirow{7}{*}{Qwen2-VL-2B-Instruct}
& Baseline & 2.9 $\pm$ 0.8 & 1.1 $\pm$ 0.4 & 0.8 $\pm$ 0.5 & 7.7 $\pm$ 3.0 & 15.8 $\pm$ 1.6 \\
& Random Image & 3.6 $\pm$ 0.8 & 0.5 $\pm$ 0.3 & 1.8 $\pm$ 0.7 & 7.7 $\pm$ 3.0 & 15.8 $\pm$ 1.6 \\
& Fixed Seed Image & 3.0 $\pm$ 0.8 & 0.7 $\pm$ 0.4 & 1.7 $\pm$ 0.7 & 8.3 $\pm$ 3.2 & 12.9 $\pm$ 1.5 \\
& Random String & 4.0 $\pm$ 0.9 & 1.9 $\pm$ 0.6 & 6.9 $\pm$ 1.3 & 30.0 $\pm$ 5.2 & 14.9 $\pm$ 1.6 \\
\cmidrule{2-7}
& Soft Prompt & \textbf{46.1} $\pm$ 2.2 & \underline{37.6} $\pm$ 2.2 & \textbf{49.7} $\pm$ 2.7 & 35.0 $\pm$ 5.3 & \textbf{77.5} $\pm$ 1.9 \\
& Optimized Image (ART) & \underline{45.0} $\pm$ 2.5\,{\scriptsize(1.5)} & \textbf{39.4} $\pm$ 2.2\,{\scriptsize(0.9)} & 40.8 $\pm$ 10.6\,{\scriptsize(11.6)} & \underline{37.4} $\pm$ 5.4\,{\scriptsize(0.4)} & 53.5 $\pm$ 3.7\,{\scriptsize(3.4)} \\
& LoRA & 15.3 $\pm$ 1.6 & 33.7 $\pm$ 2.0 & \underline{49.5} $\pm$ 2.7 & \textbf{46.7} $\pm$ 5.7 & \underline{60.1} $\pm$ 2.2 \\
\midrule
\multirow{7}{*}{Qwen3.5-0.8B}
& Baseline & 23.2 $\pm$ 1.9 & 33.9 $\pm$ 2.0 & 55.6 $\pm$ 2.7 & 67.3 $\pm$ 5.3 & 51.8 $\pm$ 2.2 \\
& Random Image & 9.4 $\pm$ 1.3 & 31.4 $\pm$ 2.0 & 49.0 $\pm$ 2.7 & 70.3 $\pm$ 5.2 & 57.3 $\pm$ 2.2 \\
& Fixed Seed Image & 1.5 $\pm$ 0.5 & 22.7 $\pm$ 1.8 & 32.8 $\pm$ 2.5 & 62.3 $\pm$ 5.5 & 54.1 $\pm$ 2.2 \\
& Random String & 21.9 $\pm$ 1.8 & 27.1 $\pm$ 1.9 & 49.8 $\pm$ 2.7 & 61.3 $\pm$ 5.7 & 50.0 $\pm$ 2.2 \\
\cmidrule{2-7}
& Soft Prompt & \underline{43.9} $\pm$ 2.0\,{\scriptsize(0.8)} & \textbf{54.6} $\pm$ 2.5\,{\scriptsize(1.8)} & \underline{56.8} $\pm$ 2.7\,{\scriptsize(1.4)} & \underline{71.7} $\pm$ 5.2\,{\scriptsize(2.7)} & \textbf{98.3} $\pm$ 0.7\,{\scriptsize(0.6)} \\
& Optimized Image (ART) & 42.1 $\pm$ 3.5\,{\scriptsize(3.1)} & \underline{52.6} $\pm$ 2.5\,{\scriptsize(1.7)} & \textbf{59.4} $\pm$ 2.9\,{\scriptsize(1.8)} & \textbf{77.1} $\pm$ 4.7\,{\scriptsize(2.2)} & 76.2 $\pm$ 7.1\,{\scriptsize(7.4)} \\
& LoRA & \textbf{47.7} $\pm$ 2.5\,{\scriptsize(1.6)} & 51.1 $\pm$ 2.3\,{\scriptsize(0.9)} & 56.2 $\pm$ 2.7\,{\scriptsize(0.9)} & 70.1 $\pm$ 4.9\,{\scriptsize(1.3)} & \underline{86.3} $\pm$ 1.7\,{\scriptsize(0.8)} \\
\midrule
\multirow{7}{*}{Qwen3.5-2B}
& Baseline & 55.2 $\pm$ 2.2 & 63.2 $\pm$ 2.1 & 73.7 $\pm$ 2.4 & \textbf{86.7} $\pm$ 3.8 & 78.3 $\pm$ 1.8 \\
& Random Image & 10.9 $\pm$ 1.3 & 14.8 $\pm$ 1.6 & 0.6 $\pm$ 0.4 & 2.0 $\pm$ 1.5 & 76.9 $\pm$ 1.8 \\
& Fixed Seed Image & 26.4 $\pm$ 1.9 & 27.6 $\pm$ 1.9 & 1.7 $\pm$ 0.7 & 7.3 $\pm$ 2.8 & 78.1 $\pm$ 1.8 \\
& Random String & 48.6 $\pm$ 2.2 & 65.1 $\pm$ 2.1 & 72.1 $\pm$ 2.5 & 82.0 $\pm$ 4.3 & 73.5 $\pm$ 1.9 \\
\cmidrule{2-7}
& Soft Prompt & 48.1 $\pm$ 2.1 & \underline{67.1} $\pm$ 2.1 & \textbf{78.5} $\pm$ 2.2 & 82.0 $\pm$ 4.2 & \textbf{90.8} $\pm$ 1.3 \\
& Optimized Image (ART) & \underline{60.4} $\pm$ 2.3\,{\scriptsize(1.2)} & 65.6 $\pm$ 1.9\,{\scriptsize(0.4)} & \underline{74.9} $\pm$ 2.6\,{\scriptsize(1.8)} & 81.8 $\pm$ 4.2\,{\scriptsize(1.3)} & 80.5 $\pm$ 1.9\,{\scriptsize(1.0)} \\
& LoRA & \textbf{64.2} $\pm$ 2.1 & \textbf{67.2} $\pm$ 2.1 & 74.6 $\pm$ 2.4 & \underline{85.0} $\pm$ 4.2 & \underline{87.4} $\pm$ 1.5 \\
\bottomrule
\end{tabular}}
\end{table*}

\paragraph{Fine-tuning comparison.} 
No method dominates. Among the three fine-tuning methods, ART is first 3 times (Qwen3.5-0.8B GSM8K and SVAMP, Qwen2-VL-2B-Instruct DROP), second 7 times, and third 10 times. 
The soft prompt ranks first on every LFM2-VL-1.6B block, usually by a wide margin, and ART trails it most on ToolMind. 
On LFM2-VL-1.6B ToolMind, both LoRA (15.7\%) and ART (8.7\%) fail to learn the task.
Appendix~\ref{sec:pairwise} tests the statistical significance of these differences with a paired bootstrap.
ART is significantly ahead of the best adapter in 2 cases (Qwen3.5-0.8B SVAMP, Qwen2-VL-2B-Instruct DROP) and within noise in 6.
The tests also confirm that ART is significantly above the text-only baseline in 18 of 20 cases.

\paragraph{Stability across training seeds.} 
Across its three training seeds, the accuracy of ART has a standard deviation of at most 3.4 points in 18 of the 20 model--task cases. 
This is close to the 0.6--2.7 points of LoRA and the soft prompt, which were also trained with three seeds on Qwen3.5-0.8B (Appendix~\ref{sec:seeds}). 
The two exceptions for ART are Qwen3.5-0.8B ToolMind (7.4 points) and Qwen2-VL-2B-Instruct GSM8K (11.6 points). 
For the latter, one of the three runs never learns the answer format (27.4\% against 47.3\% and 47.6\%).

\paragraph{Extension to larger MLLMs.} 
Appendix~\ref{sec:large_models} extends the results to three larger models, each trained with one seed. 
ART is within noise of the soft prompt on DROP, GSM8K, and SVAMP and significantly above text only in 9 of 15 cases. These include every Qwen2-VL-7B-Instruct task, by 13.3 to 55.1 points. 
On GLM-4.6V-Flash and Gemma-3-12B-it, whose text-only accuracy is already 89--93\% on the math tasks, ART is significantly above text only on GSM8K and ToolMind (1.3 to 9.7 points).

\subsection{ART versus Non-Fine-Tuned Controls}
\paragraph{Text-only baseline.}
All the fine-tuning methods improve over the text-only baseline in most of the tests (Table~\ref{tab:main_results}).
The paired tests in Appendix~\ref{sec:pairwise} verify that ART is significantly above the text-only baseline in 18 of 20 cases.
Because strict scoring requires exact answers, format errors can lower the scores on the numeric tasks. Appendix~\ref{sec:format} re-checks the gains of ART with lenient scoring on the 8 numeric cases. 
The text-only baseline improves in all of them, but ART still outperforms it in 5 of them (compared to 7 in Table~\ref{tab:main_results}).

\paragraph{Text-only prompt-engineering.}
For an extended comparison, we try to improve the text-only baseline with prompt engineering. 
Appendix~\ref{sec:text_prompts} adds comparison to one-shot prompting and GEPA prompt optimization \citep{agrawal2026gepa}.
ART is significantly ahead of the one-shot prompt in 16 of 20 cases. 
ART is significantly ahead of GEPA on GSM8K, SVAMP, and DROP (only compared on Qwen3.5-0.8B). 
GEPA is ahead of ART on ToolMind, and on bAbI it is ahead of every method.
Thus, prompt optimization remains a strong alternative to fine-tuning on selected tasks.

\paragraph{Random image.} 
Since MLLMs degrade on text-only input \citep{kim2026bridging} and random images are not neutral extra tokens \citep{darcet2024registers,goyal2024pause}, even an uninformative image can change the result.
In our runs, random images raise accuracy in some cases and lower it in others, and they slightly shorten the outputs, consistent with \citet{sun2026reading} (Appendix~\ref{sec:length}).
The largest negative effect is on the strongest baseline: a random image collapses Qwen3.5-2B GSM8K from 73.7\% to 0.6\%.
The collapse is a format failure. The model copies the answer template literally, while the last number in its output is still correct 81.4\% of the time.

\paragraph{Fixed seed image.} 
Comparing ART with its seed image isolates what the optimization changes.
Unlike the random image, the seed image does not shorten the output, whereas ART shortens it on several tests (Appendix~\ref{sec:length}).
Despite shorter reasoning, ART is significantly above the seed image in all 20 cases (Appendix~\ref{sec:pairwise}). 
\paragraph{Random string.} 
Random strings improve results over the text-only baseline on some math tasks (SVAMP for LFM2-VL-1.6B and Qwen2-VL-2B-Instruct), while they hurt others (Qwen3.5-0.8B GSM8K). 
At the same time, this control performs worse than ART fine-tuning in all cases but one. 
In the remaining case, the text-only baseline is the strongest.

\subsection{Training Cost}
\label{sec:train_cost_main}

Because the gradient of ART also passes through the vision tower, its training could be more expensive than that of the adapters.
Table~\ref{tab:train_cost} reports the mean GPU training time of each method across the five benchmarks. 
It counts only the training steps (rollouts, forward and backward passes) and leaves out model loading and the validation. 
On LFM2-VL-1.6B and Qwen2-VL-2B-Instruct the soft prompt is the cheapest. 
ART takes 1.5--1.6$\times$ longer because its gradient also passes through the vision tower. LoRA takes 2.4--2.6$\times$ the time of ART.
On Qwen3.5, ART is the fastest method (37 and 44 minutes). The soft prompt takes 1.8--2.2$\times$ and LoRA 4.3--4.4$\times$ the time of ART. The mixed architecture of Qwen3.5 is challenging to work with. Even with fast kernels for the linear-attention layers, these models are slower than their similarly sized counterparts.
A per-benchmark breakdown with training curves and timings (including larger models) is shown in Appendix~\ref{sec:training_appendix}.

\begin{table}[t]
\centering
\caption{Mean GPU training time in minutes over the five benchmarks. Times exclude model loading and validation.}
\label{tab:train_cost}
\small
\setlength{\tabcolsep}{4pt}
\begin{tabular}{lccc}
\toprule
\textbf{Model} & \textbf{ART} & \textbf{Soft prompt} & \textbf{LoRA} \\
\midrule
LFM2-VL-1.6B & 32 & \textbf{21} & 76 \\
Qwen2-VL-2B & 33 & \textbf{21} & 87 \\
Qwen3.5-0.8B & \textbf{37} & 84 & 160 \\
Qwen3.5-2B & \textbf{44} & 77 & 196 \\
\bottomrule
\end{tabular}
\end{table}

\subsection{Serving Efficiency}
\label{sec:serving_cost}

The practical case for ART, similar to soft prompts, is serving one open-weight model to many users, adapting to different tasks on the fly. 
This can be particularly useful when models must run locally because data cannot be shared externally, while still supporting different task-specific adaptations.
A single LoRA adapter can be merged into the model weights with no performance overhead. Adapting to several tasks would either require merging several LoRA adaptations into one, reducing their task-specific accuracy \citep{zhao2025lego}, or loading each adapter and applying it per request. The latter option of unmerged LoRA is used as our testing baseline.

We compare PEFTs' performance on vLLM with every request forced to decode the same number of tokens. No method can end a request early, so throughput differences reflect pure serving cost. Table~\ref{tab:serving_main} shows Qwen3.5-0.8B, GLM-4.6V-Flash, and a 4-bit quantized Qwen3.5-27B on two different GPUs. For completeness, Appendix~\ref{sec:inference_cost} reports more detailed results and extends performance benchmarking to two other MLLMs.

\paragraph{Single-task throughput.} 
ART keeps $0.86$--$0.94\times$ of the text-only throughput in every setting of Table~\ref{tab:serving_main}. Its only overhead is a one-time prefill cost from the vision tower, and its per-token decode cost stays within $6\%$ to $14\%$ of the frozen baseline. 
Like ART, a soft prompt enters the model as input embeddings and leaves the decode path untouched. However, they perform worse for the smallest 0.8B model, dropping to $0.83$--$0.88\times$. We hypothesize that this is because of the interface difference. A soft prompt is not a standard request and needs the same model-specific routing as in training (Appendix~\ref{sec:counterparts}). 
Unmerged LoRA instead pays on \emph{every} decoded token and falls to $0.26$--$0.87\times$, running slowest on V100. We note that the single-task scenario can use a merged LoRA adapter running at the baseline speed ($1.00\times$, Appendix~\ref{sec:inference_cost}).

\paragraph{Eight-task throughput.} 
Serving eight \emph{different} task adapters in one batch is free for ART and soft prompts, since eight distinct images cost the same as one image repeated, whereas LoRA's multi-adapter path, the S-LoRA setting, drops to $0.07$--$0.77\times$.

\paragraph{Server-side considerations.} 
An ART image is an ordinary PNG file, which is easy to send or share.
To adapt the target MLLM, one can simply include it in the prompt or attach it in any chat interface that accepts images. The image, however, must keep its size and its full color resolution (Section~\ref{sec:robustness}). A LoRA adapter or soft-prompt tensor, by contrast, needs a special server-side loader that accepts the possibly variable tensor-saving format. Additionally, these specific file formats are not universally supported on online sharing platforms.

\begin{table}[t]
\centering
\caption{Serving throughput relative to the text-only baseline (vLLM, batch $16$, $128$ decode tokens per request). ``$\times$ 8'' serves eight different images, soft prompts, or LoRA adapters in one batch. The models are Qwen3.5-0.8B, GLM-4.6V-Flash, and 4-bit quantized Qwen3.5-27B. 
The best results in every column are highlighted.}
\label{tab:serving_main}
\small
\setlength{\tabcolsep}{3pt}
\begin{tabular}{lcccccc}
\toprule
 & \multicolumn{2}{c}{\textbf{0.8B}} & \multicolumn{2}{c}{\textbf{GLM}} & \multicolumn{2}{c}{\textbf{27B}} \\
\cmidrule(lr){2-3}\cmidrule(lr){4-5}\cmidrule(lr){6-7}
\textbf{Method} & \textbf{A100} & \textbf{V100} & \textbf{A100} & \textbf{V100} & \textbf{A100} & \textbf{V100} \\
\midrule
ART image & \textbf{0.91} & \textbf{0.91} & 0.92 & \textbf{0.92} & \textbf{0.92} & \textbf{0.86} \\
\quad $\times$ 8 & \textbf{0.91} & 0.90 & \textbf{0.94} & \textbf{0.92} & 0.89 & \textbf{0.86} \\
Soft prompt & 0.83 & 0.88 & 0.93 & \textbf{0.92} & \textbf{0.92} & \textbf{0.86} \\
\quad $\times$ 8 & 0.83 & 0.88 & 0.93 & \textbf{0.92} & \textbf{0.92} & \textbf{0.86} \\
LoRA & 0.63 & 0.26 & 0.87 & 0.31 & 0.77 & 0.48 \\
\quad $\times$ 8 & 0.49 & 0.10 & 0.77 & 0.07 & 0.57 & 0.16 \\
\bottomrule
\end{tabular}
\end{table}

\subsection{Information Storage in ART}
\label{sec:resultsStorage}
\begin{figure}[!htb]
\centering
\includegraphics[width=0.24\linewidth]{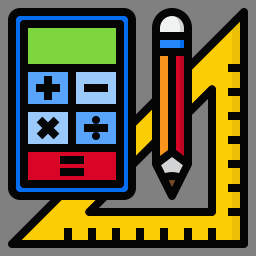}
\includegraphics[width=0.24\linewidth]{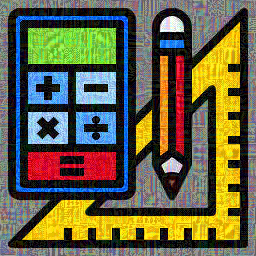}
\includegraphics[width=0.24\linewidth]{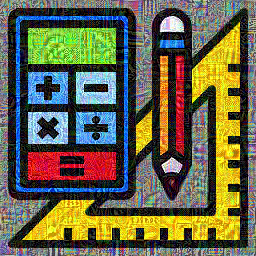}
\includegraphics[width=0.24\linewidth]{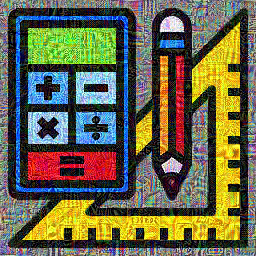}
\caption{ART images at steps 0, 60, 120, and 240 (LFM2-VL-1.6B, GSM8K, seed 1).}
\label{fig:image_progression}
\end{figure}

Since the model stays frozen, everything ART learns must be stored in the pixels of its image. 
We examine how the image changes during training and how much information it gains.
Optimization gradually overlays high-frequency structure across the entire seed image (Figure~\ref{fig:image_progression}), moving the ART image away from its seed: three quarters of all pixel values move by more than $8/255$, the usual bound for adversarial images (Appendix~\ref{sec:seed_distance}).
The per-channel entropy rises from about 1.3--1.9 bits to 7.4--7.7 bits, close to the 8-bit maximum (GSM8K image of Qwen3.5-0.8B), so the resulting 165--184~KB PNG is nearly incompressible and 5--45$\times$ larger than the seed image (Appendix~\ref{sec:pixel_stats}).

The encoding in the ART artifacts is model-specific. 
An ART image optimized for one model does not even transfer to another model of the same family (larger or smaller), often dropping test results below the text-only baseline (Appendix~\ref{sec:transfer}).
The resulting PNG images can be interpreted as constructively generated \textbf{``AI steganography''} (from the Greek ``covered writing''), with information encoded for the chosen MLLM's vision tower rather than for human interpretation.

\subsection{Robustness of ART Images to Compression}
\label{sec:robustness}

Because ART artifacts are meant to be shared as ordinary images, they may be re-encoded, resized, or compressed. 
We test the ART image robustness on every seed-1 ART image of Qwen3.5-0.8B, LFM2-VL-1.6B, and Qwen2-VL-2B-Instruct, re-running the full test evaluation. 
Specifically, we resize the $256{\times}256$ images to $512$ or $128$ pixels, and also make a round trip to $128$ or $64$ pixels and back. 
We also re-encode images as JPEG at quality $50$ to $100$, both with the default 4:2:0 chroma subsampling of PIL and libjpeg and without the subsampling (4:4:4). 
To find which step of JPEG causes the loss, we also apply its color conversion and its chroma subsampling alone. 
Finally, we convert images to a GIF-style 256-color palette.

Table~\ref{tab:robustness} summarizes the results. Mean is the mean test accuracy over the five tasks and $\Delta$ its change against the unperturbed image. The B/A column counts the tasks where the perturbed image is significantly below the unperturbed image (B) and significantly above text only (A). A is a weak bar for Qwen2-VL-2B-Instruct, whose text-only baseline is near zero (Appendix~\ref{sec:format}).

\begin{table*}[!t]
\centering
\caption{ART accuracy (\%) after perturbing the image file.}
\label{tab:robustness}
\small
\begin{tabular}{lccccccccc}
\toprule
 & \multicolumn{3}{c}{\textbf{Qwen3.5-0.8B}} & \multicolumn{3}{c}{\textbf{LFM2-VL-1.6B}} & \multicolumn{3}{c}{\textbf{Qwen2-VL-2B-Instruct}} \\
\cmidrule(lr){2-4}\cmidrule(lr){5-7}\cmidrule(lr){8-10}
\textbf{Condition} & Mean & $\Delta$ & B/A & Mean & $\Delta$ & B/A & Mean & $\Delta$ & B/A \\
\midrule
Unperturbed & 59.4 & -- & --/5 & 45.4 & -- & --/5 & 44.1 & -- & --/5 \\
Resize to 512 & 37.9 & $-$21.5 & 5/1 & 21.1 & $-$24.3 & 5/2 & 5.1 & $-$39.0 & 5/1 \\
Resize to 128 & 35.4 & $-$24.0 & 5/1 & 24.9 & $-$20.5 & 5/4 & 5.3 & $-$38.8 & 5/1 \\
128 and back & 35.5 & $-$23.9 & 5/1 & 24.1 & $-$21.3 & 5/2 & 5.8 & $-$38.3 & 5/1 \\
64 and back & 36.7 & $-$22.7 & 5/1 & 16.4 & $-$29.0 & 5/0 & 6.0 & $-$38.1 & 5/1 \\
JPEG $q{=}95$, 4:2:0 & 54.2 & $-$5.2 & 4/3 & 41.2 & $-$4.2 & 3/5 & 10.1 & $-$34.0 & 5/4 \\
JPEG $q{=}75$, 4:2:0 & 44.2 & $-$15.2 & 5/2 & 38.0 & $-$7.4 & 3/4 & 9.8 & $-$34.3 & 5/5 \\
JPEG $q{=}50$, 4:2:0 & 43.1 & $-$16.3 & 5/2 & 32.2 & $-$13.2 & 4/4 & 6.3 & $-$37.8 & 5/1 \\
JPEG $q{=}95$, 4:4:4 & 59.7 & +0.3 & 0/5 & 45.6 & +0.2 & 0/5 & 44.0 & $-$0.1 & 0/5 \\
JPEG $q{=}75$, 4:4:4 & 59.8 & +0.4 & 1/5 & 45.4 & 0.0 & 1/5 & 42.0 & $-$2.1 & 2/5 \\
4:2:0 chroma only & 55.1 & $-$4.3 & 4/3 & 40.9 & $-$4.5 & 3/4 & 10.2 & $-$33.9 & 5/4 \\
256-color palette & 58.7 & $-$0.7 & 2/5 & 44.1 & $-$1.3 & 3/4 & 38.9 & $-$5.2 & 3/5 \\
\bottomrule
\end{tabular}
\end{table*}

\paragraph{Resizing.} Resizing in either direction costs 22--24 points of mean accuracy on Qwen3.5-0.8B, 21--29 on LFM2-VL-1.6B, and 38--39 on Qwen2-VL-2B-Instruct. Every perturbed cell is significantly below the unperturbed image. The round trip to $128$ pixels and back keeps the visual-token count unchanged and still loses as much as the plain resize to $128$ ($35.5\%$ vs $35.4\%$ on Qwen3.5-0.8B).

\begin{figure*}[!t]
\centering
\includegraphics[width=\textwidth]{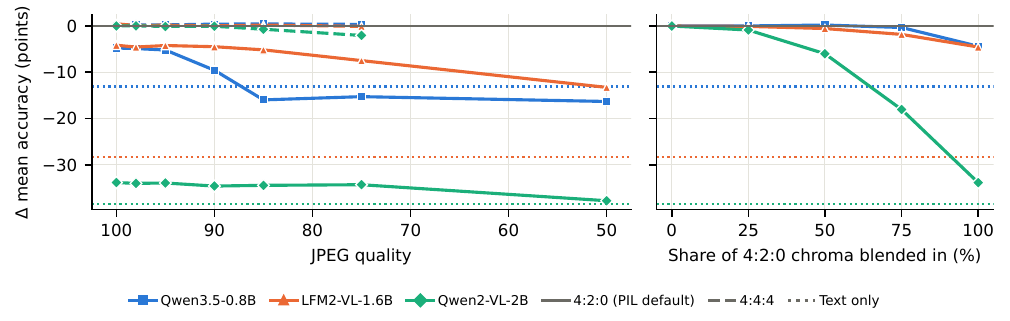}
\caption{JPEG breakdown. Change in mean test accuracy against the unperturbed image. Dotted lines mark text only.}
\label{fig:jpeg}
\end{figure*}

\paragraph{JPEG.} JPEG converts the image to YCbCr, averages the two color channels over $2{\times}2$ pixel blocks (4:2:0), and then quantizes. The color conversion alone changes mean accuracy by at most $0.3$ points. Chroma subsampling alone, saved losslessly, reproduces the whole loss of high-quality JPEG ($-4.3$, $-4.5$, and $-33.9$ points, against $-5.2$, $-4.2$, and $-34.0$ at $q{=}95$). Default JPEG loses as much at $q{=}100$. The left panel of Figure~\ref{fig:jpeg} compares default JPEG (solid) with JPEG without chroma subsampling (dashed) across quality levels. Without subsampling, JPEG is harmless. At $q{=}95$ no cell is significantly below the unperturbed image, and at $q{=}75$ the mean changes by at most $2.1$ points. For best results, ART images should therefore be shared as PNG or as JPEG without chroma subsampling.

\paragraph{Chroma sensitivity across models.} The right panel of Figure~\ref{fig:jpeg} blends the subsampled image with the original in steps of 25\%, saved losslessly, to show how the loss grows. Qwen3.5-0.8B and LFM2-VL-1.6B lose at most $1.8$ points up to a 75\% blend. Qwen2-VL-2B-Instruct loses $0.9$, $6.0$, and $18.1$ points at 25, 50, and 75\%, and full subsampling brings it close to its text-only level. Subsampling only horizontally (4:2:2 JPEG at $q{=}100$) costs the other two models $0.1$ and $1.2$ points, but still costs Qwen2-VL-2B-Instruct $33.5$.

\paragraph{Color palette.} Thanks to training-time image quantization, the information survives the $32 \to 8$-bit conversion to a PNG file. The 256-color palette costs $0.7$ and $1.3$ points on Qwen3.5-0.8B and LFM2-VL-1.6B and $5.2$ points on Qwen2-VL-2B-Instruct, and the image stays significantly above the text-only baseline in 14 of 15 cases.

\section{Limitations}

\begin{itemize}
  \item ART's effectiveness is model- and task-dependent. It is strongest on small models and procedural tasks. On tasks where the MLLM has a strong baseline (Appendices~\ref{sec:pairwise} and~\ref{sec:large_models}), the best adapter remains ahead.
  \item An ART image is tied to the model it was trained on (Appendix~\ref{sec:transfer}) and does not survive file processing. Resizing costs 21--39 points of mean accuracy and default JPEG at quality 95 costs 4--34 points, because it subsamples the color channels (Section~\ref{sec:robustness}).
\item All three methods inherit GRPO run-to-run variance, and more testing would be required to check the variance of the additional SFT approach (run with one seed in Appendix~\ref{sec:methods}). On Qwen3.5-0.8B the variance of ART is close to that of the adapters except on ToolMind, where its seeds span 14.8 points (Appendix~\ref{sec:seeds}).
\end{itemize}

\section{Conclusions}

We have presented \textbf{ART}: a non-intrusive method that adapts frozen MLLMs for text reasoning by optimizing input images. 
In our extensive comparison with established fine-tuning techniques, \textbf{no method dominated}. ART was significantly ahead of both LoRA and soft prompts in 2 of 20 cases, within noise in 6, and behind in the rest. 
ART's main advantage, however, is its computational \textbf{efficiency and serving simplicity}. An ART request is a standard multimodal image+text prompt that needs no adapter-aware infrastructure, yielding much higher throughput than unmerged LoRA typically used in multi-task serving, for example, when the model must run locally to keep data private.
Finally, with ART, we could generate visually distinctive \textbf{images styled to the task}. Sharing and using them is as simple as sending a PNG file.

\section*{Ethical Statement}

This work studies constructive uses of visual steering in MLLMs. The same visual channel mechanisms have been used for adversarial attacks. We highlight this dual use to keep practitioners aware of input-space vulnerabilities when deploying MLLMs. A shared image can change a model's behavior without being readable by eye. As with shared LoRA files, users should load images only from trusted sources.

\subsection*{AI Use Statement}

In this work, we used generative AI tools to help implement the methods (optimizing and debugging parts of the code), to help schedule and run the experiments, and to help interpret the results. We have not used generative AI tools to propose or refine the research hypotheses, to design the methods, or to clean and reformat the datasets, which the authors implemented themselves. Generating synthetic data, developing theoretical models, formulating or proving mathematical claims, translation, and qualitative data analysis are not applicable to this work. Additionally, we used generative AI tools to polish the plotting code of the figures, to search for related literature, and to draft and edit parts of the paper for readability. The tools were AI coding and writing assistants based on large language models. We have reviewed all AI-assisted work. All AI-assisted code was reviewed by the authors and run for every reported experiment. Every number in the tables and the text was recomputed from the raw result files by scripts and checked by the authors, and every AI-suggested reference was checked against the original paper. We take responsibility for the final content of this work, including text, claims or artifacts produced with the aid of generative AI.

\section*{Acknowledgments}
This work is part of the Center for Research-based Innovation DigiWells, which stands for Digital Well Center for Value Creation, Competitiveness and Minimum Environmental Footprint (NFR SFI project no. 309589, \url{https://DigiWells.no}). The center is a cooperation of NORCE Norwegian Research Centre, the University of Stavanger, the Norwegian University of Science and Technology (NTNU), and the University of Bergen. It is funded by Aker BP, ConocoPhillips, Equinor, Harbour Energy, Petrobras, TotalEnergies, Vår Energi, and the Research Council of Norway.

\bibliography{references}

\let\origsection\section
\renewcommand{\section}{\FloatBarrier\origsection}
\clearpage
\appendix
\begin{center}
{\Large\bfseries Appendices\par}
\end{center}

\section{Benchmark Datasets}
\label{sec:datasets_detail}

Evaluation uses strict binary scoring on every benchmark. Each entry below describes the scorer and the shaped training reward (motivation in Section~\ref{sec:datasets}).

\paragraph{GSM8K (Grade-School Math).}
GSM8K \citep{gsm8k} consists of grade-school math word problems that require multi-step arithmetic.
The scorer extracts the final numeric value and checks for an exact match.
The training reward adds partial credit for correctly formatted but numerically near answers.

\paragraph{SVAMP (Math Word Problems).}
SVAMP \citep{patel2021svamp} extends grade-school-style math word problems with variations in number order, sentence structure, and problem composition.
Scoring and training reward are the same as for GSM8K.

\paragraph{DROP (Reading Comprehension).}
DROP \citep{dua2019drop} is a reading-comprehension benchmark that requires discrete reasoning over paragraphs, including counting, sorting, and arithmetic on numbers extracted from the passage.
The scorer performs a strict span match against the ground-truth answer.
The training reward is a token-level F1 over the span.

\paragraph{bAbI (Synthetic Reasoning).}
bAbI \citep{weston2015babi} is a suite of synthetic reasoning tasks that test basic inference such as deduction, induction, and path finding over short stories.
The scorer is a strict answer match against the gold answer.
The training reward adds a small format floor for a parseable answer.

\paragraph{ToolMind (Structured Tool Use).}
ToolMind \citep{nanbeige2025toolmind} evaluates an LLM's capacity to construct structured XML-form API function calls based on environmental parameters.
We process the first user-to-assistant turn.
The scorer extracts the generated function name and arguments and performs a strict match against the ground-truth parameters, scoring $1$ if the correct function is called and all required arguments are present, and $0$ otherwise.
The training reward is $0.3\,m + 0.7\,f$, where $m$ indicates a matched tool call and $f$ the fraction of correctly filled arguments.

\section{Additional Training Objective}
\label{sec:methods}

This appendix tests supervised fine-tuning (SFT) \citep{vonwerra2022trl} as an alternative training method for producing ART images. Both training stratetegies produce ART images that improve MLLMs.

\begin{table}[!htb]
\centering
\caption{GSM8K accuracy (\%) for alternative image-training objectives. Best in \textbf{bold}.}
\label{tab:methods}
\small
\setlength{\tabcolsep}{2pt}
\begin{tabular}{lcc}
\toprule
\textbf{Objective} & \textbf{LFM2-VL-1.6B} & \textbf{Qwen2-VL-2B-Instruct} \\
\midrule
SFT Image & 55.2 $\pm$ 2.7 & 45.8 $\pm$ 2.7 \\
GRPO Image (ART) & \textbf{60.0} $\pm$ 2.7 & \textbf{47.3} $\pm$ 2.7 \\
\bottomrule
\end{tabular}
\end{table}

\begin{figure*}[!t]
\centering
\includegraphics[width=0.8\textwidth]{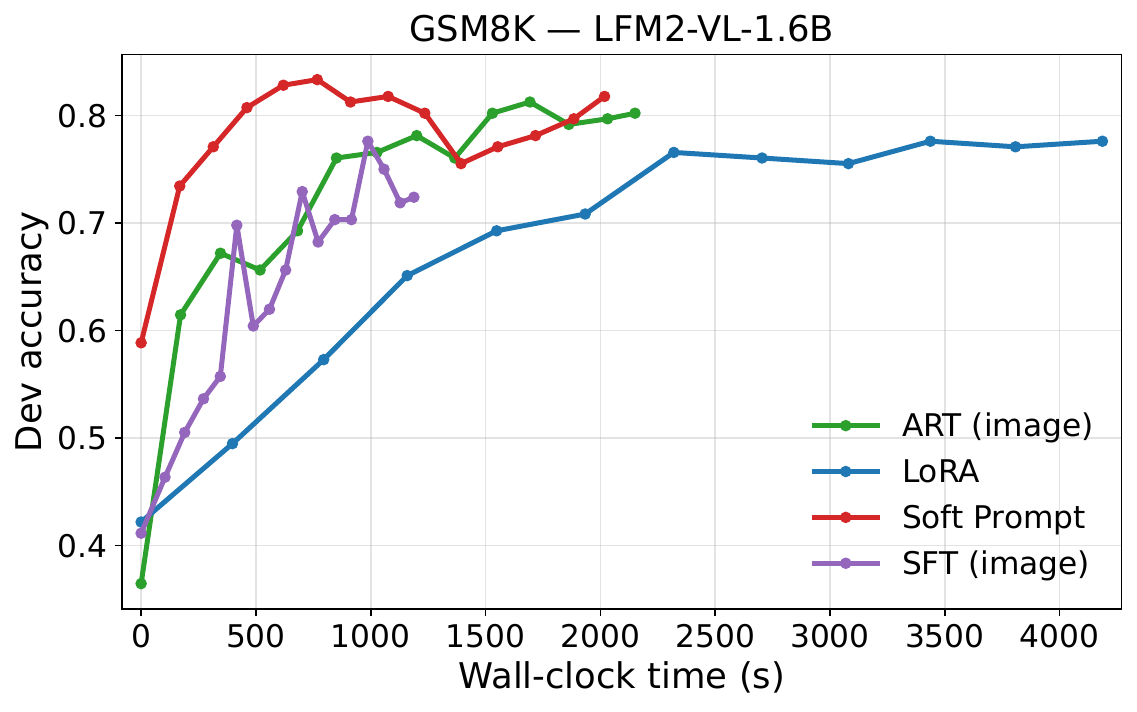}
\caption{Development accuracy on GSM8K against wall-clock time for LFM2-VL-1.6B. Image objectives (SFT, GRPO/ART) are shown with LoRA and the soft prompt.}
\label{fig:full_acc}
\end{figure*}

Table~\ref{tab:methods} compares GRPO with supervised fine-tuning (SFT) for image optimization on GSM8K, on two MLLMs. Both objectives use one run with seed 1.
GRPO yields the highest accuracy on both models. On LFM2-VL-1.6B, SFT reaches 55.2\% against 60.0\% for GRPO, and Figure~\ref{fig:full_acc} shows that it climbs quickly but noisily and stays below the GRPO image. On Qwen2-VL-2B-Instruct, SFT (45.8\%) is close to GRPO (47.3\%).

\section{LoRA and Soft Prompts: Methodology and Implementation}
\label{sec:counterparts}

LoRA is implemented with the Hugging Face PEFT library \citep{mangrulkar2022peft}, with the adapter target modules scoped to the language-model projections only so the vision tower stays frozen.

Soft prompts occupy an intermediate position between LoRA and ART. They are input-space prefixes like ART, but they live in embedding space and bypass the vision tower.
Where the soft prompt leads, its edge fits its greater freedom, since it writes arbitrary embedding vectors while ART is confined to the vision tower's image manifold and 8-bit quantization. 
Structured tasks such as ToolMind, which reward a well-formed call, favor the embedding prefix most.

Importantly, soft prompts still require a tensor input path, whereas ART's artifact is a plain image. 
This path made soft prompting the most challenging of the three methods to train and evaluate under GRPO. 
The \texttt{prompt\_embeds} path of vLLM 0.19.1, which accepts precomputed embeddings instead of token IDs, exhibited model-specific issues that forced us to split the routing in two by positional-encoding scheme. 
M-RoPE models (e.g.\ Qwen3.5) are routed through the \texttt{image\_embeds} input with manually computed position IDs, while plain-RoPE models (e.g.\ LFM2-VL) need a \texttt{language\_model\_only} flag so the vision tower does not corrupt the embedded prefix.

\section{Training Config}
\label{sec:config}

\begin{table}[!htb]
\centering
\caption{Training configuration.}
\label{tab:config}
\small
\begin{tabular}{ll}
\toprule
\textbf{Setting} & \textbf{Value} \\
\midrule
ART image & Lion ($\beta_1{=}\beta_2{=}0.95$), LR $0.003$, weight decay $0$ \\
Soft prompt & AdamW, LR $0.01$, weight decay $0$ \\
LoRA & AdamW, LR $10^{-5}$, weight decay $0.01$, $r{=}8$, $\alpha{=}8$, dropout $0$ \\
Group size & $32$ \\
Batch & $4$ per device, accumulation $16$ (effective $64$ rollouts) \\
Rollout sampling & temperature $0.85$, min-$p$ $0.01$ \\
LR schedule & cosine, $10\%$ warmup \\
Image size & $256{\times}256$ \\
Max completion & $384$ tokens \\
Data & one epoch over the first $700$ examples, last $192$ as validation probe \\
\bottomrule
\end{tabular}
\end{table}

Table~\ref{tab:config} lists the training configuration shared by the three methods. The loss is Dr.~GRPO. Advantages are mean-centered without the standard-deviation denominator, truncated completions are masked, and each rollout is consumed by exactly one gradient step. 
All three methods select their checkpoint with the same probe, greedy decoding, strict scorer, and rule (the latest checkpoint within a standard error of the best probe accuracy). LoRA updates the query, key, value, output, gate, up, and down projections of the language model, with rank and $\alpha$ following the Unsloth LoRA hyperparameter guide~\citep{unsloth2024lora} for reinforcement-learning fine-tuning. Hyperparameters were set once and shared across benchmarks. GSM8K and SVAMP start from the math seed image, DROP and bAbI from the brain seed, and ToolMind from the tool seed (Figure~\ref{fig:optimized_images}).

\begin{table}[!htb]
\centering
\caption{Trainable parameters per method.}
\label{tab:params}
\small
\begin{tabular}{lrrrr}
\toprule
\textbf{Model} & \textbf{Frozen} & \textbf{ART} & \textbf{Soft} & \textbf{LoRA} \\
\midrule
Qwen3.5-0.8B & 853M & 197K & 66K & 3.19M \\
Qwen2-VL-2B & 2.21B & 197K & 124K & 9.23M \\
LFM2-VL-1.6B & 1.58B & 197K & 131K & 5.55M \\
Qwen3.5-2B & 2.21B & 197K & 131K & 5.46M \\
\bottomrule
\end{tabular}
\end{table}

Table~\ref{tab:params} lists the trainable parameters of each method, as logged at the start of training.
The ART image always has $3{\times}256{\times}256 = 196{,}608$ values, independent of the model.
With one embedding row per visual token (64 or 81 rows times the hidden size), the soft prompt is 1.5 to 3 times smaller than the image.
Since a pixel takes one byte and a bfloat16 value two, the largest possible artifact is 0.1875\,MB for ART, 0.125--0.25\,MB for the soft prompt, and 6.1--17.6\,MB for LoRA.

\section{Statistical Significance of Performance Changes over Baselines}
\label{sec:pairwise}

\begin{table*}[!t]
\centering
\caption{
  Paired differences (percentage points) with 95\% paired-bootstrap intervals, computed from the same runs as in Table~\ref{tab:main_results}.
  `Best' is the stronger of LoRA and the soft prompt (`Soft') for that task.
  Statistically significant differences, whose interval excludes zero, are in \textbf{bold}.
  Significant improvements, whose interval lies above zero, are also \underline{underlined}.
  }
\label{tab:paired1}
\small
\setlength{\tabcolsep}{4pt}
\resizebox{\textwidth}{!}{\begin{tabular}{lllr@{\hspace{3pt}}r@{\hspace{2pt}}rr@{\hspace{3pt}}r@{\hspace{2pt}}rr@{\hspace{3pt}}r@{\hspace{2pt}}r}
\toprule
\textbf{Model} & \textbf{Task} & \textbf{Best} & \multicolumn{3}{c}{\textbf{ART} minus \textbf{`Best'}} & \multicolumn{3}{c}{\textbf{ART} minus \textbf{seed image}} & \multicolumn{3}{c}{\textbf{ART} minus \textbf{text only}} \\
\midrule
LFM2-VL-1.6B & bAbI & Soft & \textbf{$-$11.8} & \textbf{[$-$14.2,} & \textbf{$-$9.5]} & \textbf{\underline{+34.9}} & \textbf{[+32.5,} & \textbf{+37.2]} & \textbf{\underline{+37.6}} & \textbf{[+35.1,} & \textbf{+39.9]} \\
 & DROP & Soft & \textbf{$-$12.1} & \textbf{[$-$14.7,} & \textbf{$-$9.7]} & \textbf{\underline{+17.4}} & \textbf{[+15.0,} & \textbf{+19.8]} & \textbf{\underline{+19.7}} & \textbf{[+17.2,} & \textbf{+22.2]} \\
 & GSM8K & Soft & \textbf{$-$4.7} & \textbf{[$-$7.3,} & \textbf{$-$2.3]} & \textbf{\underline{+41.4}} & \textbf{[+38.4,} & \textbf{+44.4]} & \textbf{\underline{+39.0}} & \textbf{[+35.9,} & \textbf{+41.9]} \\
 & SVAMP & Soft & $-$4.3 & [$-$8.9, & +0.3] & \textbf{\underline{+51.7}} & \textbf{[+45.9,} & \textbf{+57.4]} & \textbf{\underline{+39.0}} & \textbf{[+33.2,} & \textbf{+45.0]} \\
 & ToolMind & Soft & \textbf{$-$85.5} & \textbf{[$-$87.4,} & \textbf{$-$83.6]} & \textbf{\underline{+2.5}} & \textbf{[+1.1,} & \textbf{+4.1]} & \textbf{\underline{+2.2}} & \textbf{[+0.6,} & \textbf{+3.9]} \\
\midrule
Qwen2-VL-2B-Instruct & bAbI & Soft & $-$1.1 & [$-$3.2, & +1.2] & \textbf{\underline{+42.0}} & \textbf{[+39.5,} & \textbf{+44.5]} & \textbf{\underline{+42.1}} & \textbf{[+39.5,} & \textbf{+44.7]} \\
 & DROP & Soft & \textbf{\underline{+1.8}} & \textbf{[+0.1,} & \textbf{+3.5]} & \textbf{\underline{+38.7}} & \textbf{[+36.6,} & \textbf{+41.0]} & \textbf{\underline{+38.3}} & \textbf{[+36.1,} & \textbf{+40.6]} \\
 & GSM8K & Soft & \textbf{$-$8.9} & \textbf{[$-$21.5,} & \textbf{$-$0.4]} & \textbf{\underline{+39.1}} & \textbf{[+26.4,} & \textbf{+47.5]} & \textbf{\underline{+40.0}} & \textbf{[+27.1,} & \textbf{+48.4]} \\
 & SVAMP & LoRA & \textbf{$-$9.3} & \textbf{[$-$14.3,} & \textbf{$-$4.1]} & \textbf{\underline{+29.1}} & \textbf{[+24.0,} & \textbf{+34.3]} & \textbf{\underline{+29.7}} & \textbf{[+24.7,} & \textbf{+35.1]} \\
 & ToolMind & Soft & \textbf{$-$24.0} & \textbf{[$-$27.7,} & \textbf{$-$20.6]} & \textbf{\underline{+40.6}} & \textbf{[+36.7,} & \textbf{+44.0]} & \textbf{\underline{+37.7}} & \textbf{[+33.8,} & \textbf{+41.1]} \\
\midrule
Qwen3.5-0.8B & bAbI & LoRA & \textbf{$-$5.6} & \textbf{[$-$8.9,} & \textbf{$-$1.9]} & \textbf{\underline{+40.6}} & \textbf{[+37.3,} & \textbf{+44.3]} & \textbf{\underline{+18.9}} & \textbf{[+15.8,} & \textbf{+22.5]} \\
 & DROP & Soft & $-$2.0 & [$-$4.9, & +0.9] & \textbf{\underline{+29.9}} & \textbf{[+27.3,} & \textbf{+32.5]} & \textbf{\underline{+18.7}} & \textbf{[+16.1,} & \textbf{+21.4]} \\
 & GSM8K & Soft & +2.6 & [$-$0.5, & +5.6] & \textbf{\underline{+26.6}} & \textbf{[+23.3,} & \textbf{+29.7]} & \textbf{\underline{+3.8}} & \textbf{[+0.7,} & \textbf{+6.8]} \\
 & SVAMP & Soft & \textbf{\underline{+5.4}} & \textbf{[+0.2,} & \textbf{+11.0]} & \textbf{\underline{+14.8}} & \textbf{[+9.3,} & \textbf{+20.4]} & \textbf{\underline{+9.8}} & \textbf{[+4.1,} & \textbf{+15.6]} \\
 & ToolMind & Soft & \textbf{$-$22.1} & \textbf{[$-$29.6,} & \textbf{$-$15.5]} & \textbf{\underline{+22.1}} & \textbf{[+14.6,} & \textbf{+28.9]} & \textbf{\underline{+24.4}} & \textbf{[+17.1,} & \textbf{+31.2]} \\
\midrule
Qwen3.5-2B & bAbI & LoRA & \textbf{$-$3.8} & \textbf{[$-$5.9,} & \textbf{$-$2.0]} & \textbf{\underline{+34.0}} & \textbf{[+31.6,} & \textbf{+36.4]} & \textbf{\underline{+5.2}} & \textbf{[+3.0,} & \textbf{+7.2]} \\
 & DROP & LoRA & $-$1.6 & [$-$3.5, & +0.1] & \textbf{\underline{+38.0}} & \textbf{[+35.7,} & \textbf{+40.2]} & \textbf{\underline{+2.4}} & \textbf{[+0.4,} & \textbf{+4.2]} \\
 & GSM8K & Soft & \textbf{$-$3.6} & \textbf{[$-$6.2,} & \textbf{$-$1.0]} & \textbf{\underline{+73.2}} & \textbf{[+70.5,} & \textbf{+75.9]} & +1.2 & [$-$1.4, & +4.1] \\
 & SVAMP & LoRA & $-$3.2 & [$-$7.3, & +0.9] & \textbf{\underline{+74.5}} & \textbf{[+69.6,} & \textbf{+79.2]} & \textbf{$-$4.9} & \textbf{[$-$8.7,} & \textbf{$-$1.2]} \\
 & ToolMind & Soft & \textbf{$-$10.3} & \textbf{[$-$12.1,} & \textbf{$-$8.7]} & \textbf{\underline{+2.4}} & \textbf{[+0.9,} & \textbf{+3.6]} & \textbf{\underline{+2.2}} & \textbf{[+0.6,} & \textbf{+3.7]} \\
\bottomrule
\end{tabular}}
\end{table*}

\paragraph{Bootstrap comparison.} 
Because all methods are evaluated on the same test examples, we compare them with a paired bootstrap ($10{,}000$ resamples). 
For a three-seed method, each resample first draws the seeds with replacement, then draws the shared test examples. The interval thus covers both training and evaluation noise (Table~\ref{tab:paired1}). 
For the other models, the adapters run once, and the interval covers only their evaluation noise.

\paragraph{Ranking.} ART leads in three cases and is significantly ahead in two of them (Qwen3.5-0.8B SVAMP, $+5.4$, and Qwen2-VL-2B-Instruct DROP, $+1.8$). The third lead, Qwen3.5-0.8B GSM8K ($+2.6$), is within noise, as are five more cases (Qwen3.5-0.8B DROP, Qwen3.5-2B DROP and SVAMP, Qwen2-VL-2B-Instruct bAbI, and LFM2-VL-1.6B SVAMP). ART is significantly behind in the other 12 cases, which include every ToolMind cell and three of the four bAbI cases.

\paragraph{Seed image and text only.} ART is significantly above its own unoptimized seed image in all 20 cases and above the text-only baseline in 18. The two exceptions are both on Qwen3.5-2B. On GSM8K ART is within noise of the baseline ($74.9\%$ vs $73.7\%$), and on SVAMP it is $4.9$ points below a strong text-only baseline.

\section{Variation across Training Seeds}
\label{sec:seeds}

We compare run-to-run variance on Qwen3.5-0.8B, the only model where all three methods have three training seeds (Section~\ref{sec:experiments}). ART always starts from the same seed image.

\paragraph{Variance by method.} Table~\ref{tab:seeds} gives the standard deviation and range of the test accuracy over seeds. On bAbI, DROP, GSM8K, and SVAMP the standard deviation of ART is 1.7--3.1 points, against 0.8--2.7 for the soft prompt and 0.9--1.6 for LoRA. ART is close to the soft prompt on DROP, GSM8K, and SVAMP and the most variable method on bAbI. ToolMind is the exception. Its ART seeds span 68.4\% to 83.2\% (standard deviation 7.4), while both adapters stay within 1.6 points. On the other three models the ART standard deviation is at most 3.4 points, except for the failed Qwen2-VL-2B-Instruct GSM8K seed (Section~\ref{sec:main_results}).

\begin{table}[!htb]
\centering
\caption{Test-accuracy spread (points) over three training seeds on Qwen3.5-0.8B. Range is the highest minus the lowest seed.}
\label{tab:seeds}
\small
\setlength{\tabcolsep}{3pt}
\begin{tabular}{lcccccc}
\toprule
 & \multicolumn{2}{c}{\textbf{Soft}} & \multicolumn{2}{c}{\textbf{ART}} & \multicolumn{2}{c}{\textbf{LoRA}} \\
\cmidrule(lr){2-3}\cmidrule(lr){4-5}\cmidrule(lr){6-7}
Task & SD & Range & SD & Range & SD & Range \\
\midrule
bAbI & 0.8 & 1.5 & 3.1 & 6.0 & 1.6 & 2.9 \\
DROP & 1.8 & 3.1 & 1.7 & 3.2 & 0.9 & 1.6 \\
GSM8K & 1.4 & 2.4 & 1.8 & 3.3 & 0.9 & 1.5 \\
SVAMP & 2.7 & 5.3 & 2.2 & 4.0 & 1.3 & 2.3 \\
ToolMind & 0.6 & 1.3 & 7.4 & 14.8 & 0.8 & 1.6 \\
\bottomrule
\end{tabular}
\end{table}

\begin{figure*}[!t]
\centering
\includegraphics[width=0.48\textwidth]{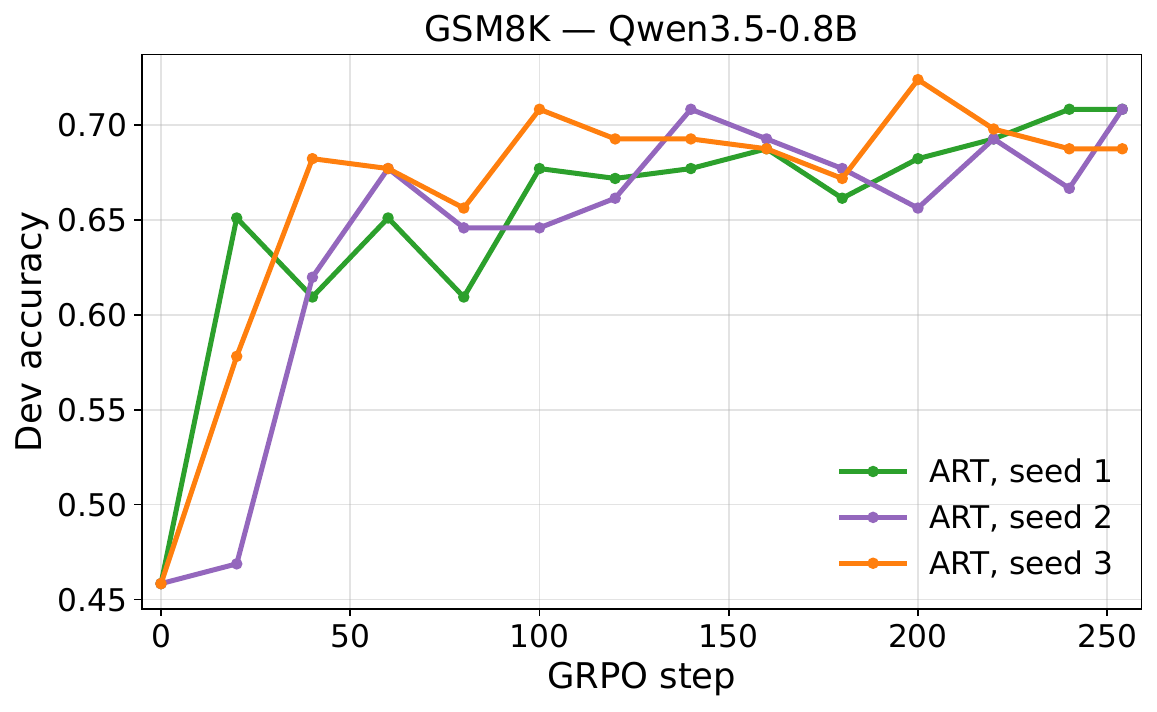}
\includegraphics[width=0.48\textwidth]{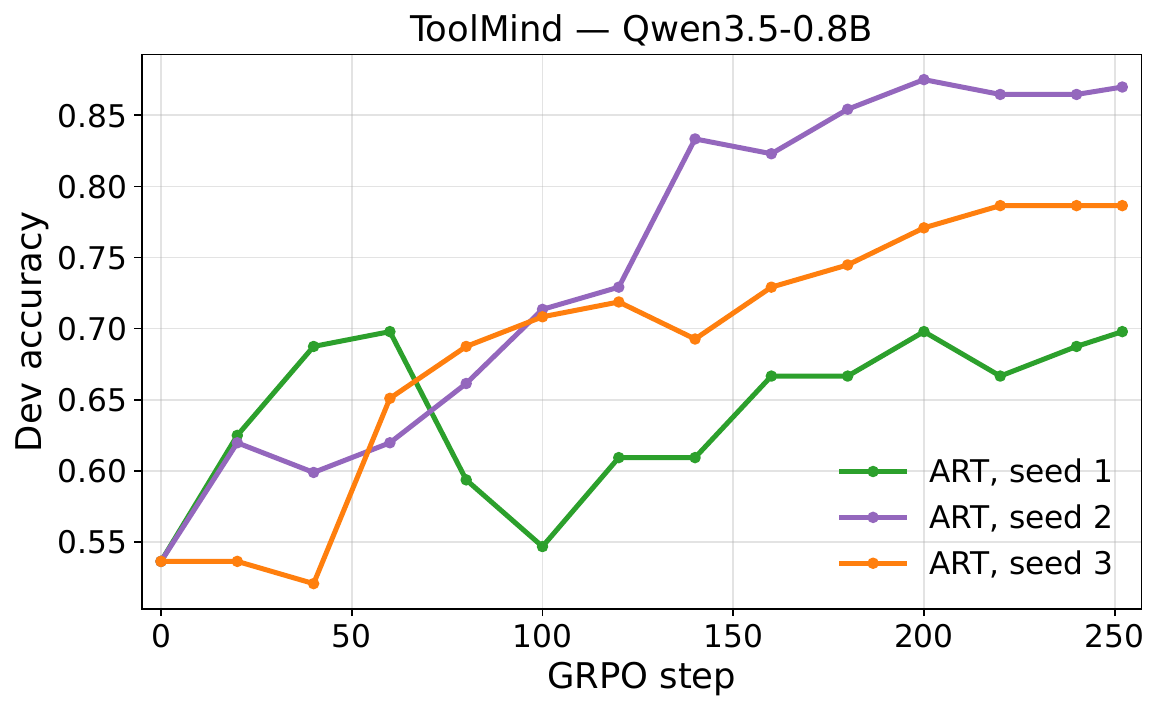}
\caption{Development accuracy of ART for three training seeds on Qwen3.5-0.8B, GSM8K (left) and ToolMind (right).}
\label{fig:seed_progression}
\end{figure*}

\paragraph{ToolMind.} For ART the seed changes only the prompt order and the rollouts sampled at temperature 0.85, and every run evaluates greedily. Figure~\ref{fig:seed_progression} shows the development accuracy. On GSM8K the seeds end within three points of each other. On ToolMind they separate after about 100 steps and end at 69.8\%, 87.0\%, and 78.6\%, matching the larger test spread in Table~\ref{tab:seeds}.

\begin{figure*}[!t]
\centering
\includegraphics[width=0.8\textwidth]{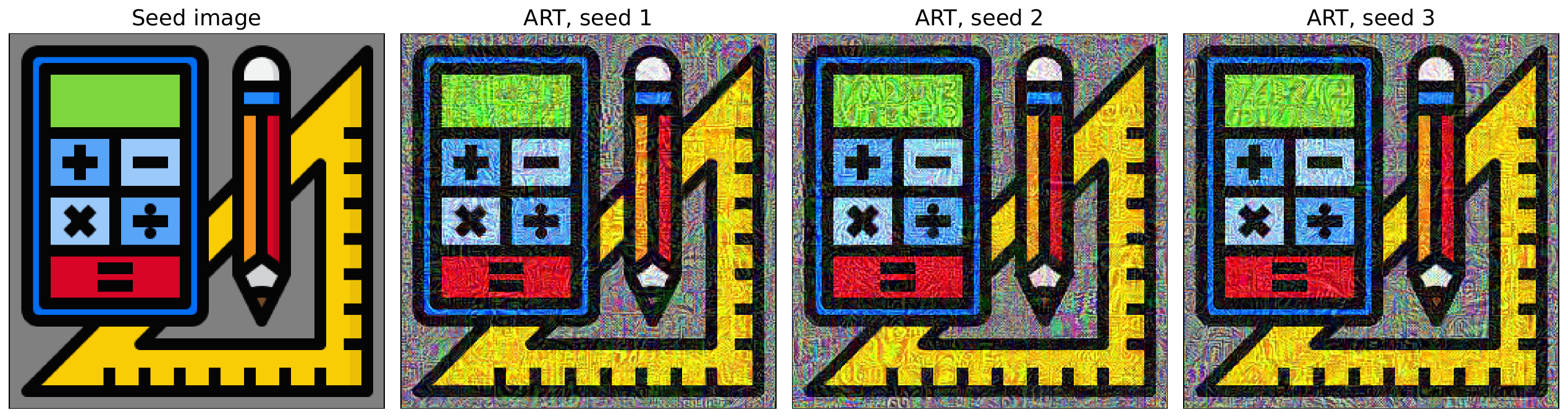}
\caption{GSM8K seed image and the ART images of the three training seeds on Qwen3.5-0.8B.}
\label{fig:seed_images}
\end{figure*}

The seeds also end on different images (Figure~\ref{fig:seed_images}). They look alike, but any two are 40--58 levels apart (RMS), more than their 33--41 levels from the seed image. Similar accuracy is therefore reached through largely different pixel codes.

\section{Larger Models}
\label{sec:large_models}

Table~\ref{tab:large_models} reports ART and the soft prompt on three larger models, each trained with one seed. The soft prompt has as many tokens as the image, 81 on Qwen2-VL-7B-Instruct and GLM-4.6V-Flash and 256 on Gemma-3-12B-it. Significance uses the paired bootstrap of Appendix~\ref{sec:pairwise}.

\begin{table*}[!t]
\centering
\caption{Accuracy (\%) of the larger models (one training seed) with 95\% bootstrap CI half-widths. Best per model in \textbf{bold}.}
\label{tab:large_models}
\small
\setlength{\tabcolsep}{4pt}
\resizebox{\textwidth}{!}{\begin{tabular}{llccccc}
\toprule
\textbf{Model} & \textbf{Condition} & \textbf{bAbI} & \textbf{DROP} & \textbf{GSM8K} & \textbf{SVAMP} & \textbf{ToolMind} \\
\midrule
\multirow{3}{*}{Qwen2-VL-7B-Instruct}
& Baseline & 3.2 $\pm$ 0.8 & 5.9 $\pm$ 1.0 & 65.3 $\pm$ 2.6 & 42.7 $\pm$ 5.5 & 54.1 $\pm$ 2.2 \\
\cmidrule{2-7}
& Soft Prompt & \textbf{65.2} $\pm$ 2.1 & \textbf{61.8} $\pm$ 2.1 & \textbf{80.7} $\pm$ 2.1 & \textbf{86.7} $\pm$ 3.8 & \textbf{82.4} $\pm$ 1.6 \\
& Optimized Image (ART) & 57.5 $\pm$ 2.2 & 61.0 $\pm$ 2.1 & 78.6 $\pm$ 2.2 & 85.0 $\pm$ 4.2 & 75.2 $\pm$ 1.9 \\
\midrule
\multirow{3}{*}{GLM-4.6V-Flash}
& Baseline & 64.0 $\pm$ 2.1 & 68.8 $\pm$ 2.0 & 89.0 $\pm$ 1.7 & 92.7 $\pm$ 2.8 & 64.3 $\pm$ 2.1 \\
\cmidrule{2-7}
& Soft Prompt & \textbf{68.2} $\pm$ 2.0 & \textbf{69.8} $\pm$ 2.0 & \textbf{91.3} $\pm$ 1.5 & \textbf{94.7} $\pm$ 2.5 & \textbf{82.6} $\pm$ 1.7 \\
& Optimized Image (ART) & 64.5 $\pm$ 2.1 & 69.0 $\pm$ 2.1 & 91.2 $\pm$ 1.6 & 92.7 $\pm$ 2.8 & 74.0 $\pm$ 1.9 \\
\midrule
\multirow{3}{*}{Gemma-3-12B-it}
& Baseline & \textbf{70.0} $\pm$ 2.0 & 72.2 $\pm$ 2.0 & 91.0 $\pm$ 1.6 & 89.3 $\pm$ 3.5 & 77.9 $\pm$ 1.8 \\
\cmidrule{2-7}
& Soft Prompt & 69.9 $\pm$ 2.0 & \textbf{74.3} $\pm$ 1.9 & 91.3 $\pm$ 1.5 & 90.0 $\pm$ 3.2 & \textbf{84.0} $\pm$ 1.6 \\
& Optimized Image (ART) & 68.1 $\pm$ 2.0 & 72.9 $\pm$ 2.0 & \textbf{92.3} $\pm$ 1.4 & \textbf{90.3} $\pm$ 3.3 & 82.6 $\pm$ 1.7 \\
\bottomrule
\end{tabular}}
\end{table*}

\paragraph{Soft prompt.} ART is within noise of the soft prompt on DROP, GSM8K, and SVAMP for all three models, 9 of the 15 cases. On Gemma-3-12B-it ART is ahead on GSM8K ($92.3\%$ vs $91.3\%$) and SVAMP ($90.3\%$ vs $90.0\%$). The soft prompt is significantly ahead on bAbI and ToolMind for all three models, by 1.4 to 8.6 points.

\paragraph{Text only.} ART is significantly above text only in 9 of 15 cases. On Qwen2-VL-7B-Instruct it gains on every task, by 13.3 to 55.1 points. On GLM-4.6V-Flash and Gemma-3-12B-it the significant gains are on GSM8K ($+2.2$ and $+1.3$) and ToolMind ($+9.7$ and $+4.7$). On their other tasks text only already reaches 64--93\%, and ART is within noise of it except on Gemma-3-12B-it bAbI, where it is 1.9 points below.

\section{Answer Format, Output Length, and Correctness}
\label{sec:format}

Strict exact match rewards both writing the answer in the required \texttt{Answer:} format and getting it right. To separate them we re-score the logged outputs of the numeric tasks in two ways (Table~\ref{tab:format}). The parse rate is the share of outputs in which the strict scorer finds a number after \texttt{Answer:}. The lenient accuracy takes the last number anywhere in the output, which can also credit an intermediate number.

\paragraph{Weak-format models.} LFM2-VL-1.6B and Qwen2-VL-2B-Instruct emit a parseable answer in only 3--42\% of their text-only outputs. Their lenient baseline accuracy is far above the strict one (63.5\% vs 21.0\% for LFM2-VL-1.6B on GSM8K), and on LFM2-VL-1.6B it is already level with all three fine-tuned methods (62.9--64.6\%). On these models all three methods mostly teach the answer format. Qwen2-VL-2B-Instruct on GSM8K still gains 8.3 points beyond format (39.3\% lenient baseline vs 47.6\% for ART).

\paragraph{Format-following models.} The Qwen3.5 models parse 88--95\% of their text-only outputs. Here ART raises lenient GSM8K accuracy by 4.6 points on Qwen3.5-0.8B (56.0\% to 60.6\%) and by 3.8 points on Qwen3.5-2B (74.1\% to 77.9\%). On Qwen3.5-2B SVAMP, ART is below the text-only baseline under both scorings.

\paragraph{Collapses under random and seed images.} Qwen3.5-2B collapses under a random or seed image (GSM8K 0.6\% and 1.7\%) because it copies the answer template literally, for example \texttt{Answer: <number>18</number>}, which the strict scorer rejects. Its lenient accuracy stays at 81.4\% and 78.1\%.

\paragraph{Output length.}\label{sec:length} 
Because image input can shorten the chain of thought of an MLLM \citep{sun2026reading}, part of a gain could also come from a change in the length of the reasoning, which we measure by the output length.
Table~\ref{tab:length} reports the mean output length relative to text only, as a geometric mean of the per-benchmark ratios over the test outputs of Table~\ref{tab:main_results}.
Averaged over the five benchmarks, ART's outputs are never longer than those of its seed image (0.69--1.00 vs 0.99--1.11 of the text-only length), yet ART is significantly more accurate in all 20 cases (Appendix~\ref{sec:pairwise}), so its gains do not come from longer reasoning.

\begin{table}[!htb]
\centering
\caption{Mean output length relative to text only over the five benchmarks. R.\ str.\ and R.\ img.\ are the random-string and random-image controls, and Seed is the unoptimized seed image.}
\label{tab:length}
\small
\setlength{\tabcolsep}{3pt}
\begin{tabular}{lcccccc}
\toprule
Model & R.\ str. & R.\ img. & Seed & ART & Soft & LoRA \\
\midrule
Qwen3.5-0.8B & 0.87 & 0.99 & 1.11 & 0.82 & 1.27 & 0.69 \\
Qwen3.5-2B & 1.03 & 0.99 & 1.01 & 1.00 & 1.19 & 0.97 \\
LFM2-VL-1.6B & 1.03 & 0.92 & 1.07 & 0.69 & 0.53 & 0.83 \\
Qwen2-VL-2B-Instr. & 0.66 & 0.84 & 0.99 & 0.74 & 0.78 & 1.19 \\
\bottomrule
\end{tabular}
\end{table}

\begin{table*}[!htb]
\centering
\caption{Strict accuracy (Acc), parse rate (Parse), and lenient accuracy (Len.) in \% on the numeric tasks, with seed means as in Table~\ref{tab:main_results}.}
\label{tab:format}
\small
\setlength{\tabcolsep}{3pt}
\begin{tabular}{llcccccccccccc}
\toprule
 & & \multicolumn{3}{c}{\textbf{Text only}} & \multicolumn{3}{c}{\textbf{ART}} & \multicolumn{3}{c}{\textbf{Soft prompt}} & \multicolumn{3}{c}{\textbf{LoRA}} \\
\cmidrule(lr){3-5}\cmidrule(lr){6-8}\cmidrule(lr){9-11}\cmidrule(lr){12-14}
\textbf{Model} & \textbf{Task} & Acc & Parse & Len. & Acc & Parse & Len. & Acc & Parse & Len. & Acc & Parse & Len. \\
\midrule
LFM2-VL-1.6B & GSM8K & 21.0 & 30.4 & 63.5 & 60.0 & 92.2 & 62.9 & 64.7 & 98.5 & 64.6 & 60.5 & 91.8 & 64.5 \\
 & SVAMP & 34.0 & 41.7 & 75.0 & 73.0 & 96.8 & 74.0 & 77.3 & 98.7 & 77.3 & 74.0 & 96.0 & 76.0 \\
\midrule
Qwen2-VL-2B-Instruct & GSM8K & 0.8 & 3.0 & 39.3 & 40.8 & 81.8 & 47.6 & 49.7 & 97.0 & 50.3 & 49.5 & 95.2 & 51.5 \\
 & SVAMP & 7.7 & 17.7 & 28.3 & 37.4 & 100.0 & 37.4 & 35.0 & 100.0 & 35.0 & 46.7 & 97.3 & 49.7 \\
\midrule
Qwen3.5-0.8B & GSM8K & 55.6 & 87.8 & 56.0 & 59.4 & 88.3 & 60.6 & 56.8 & 84.4 & 58.5 & 56.2 & 90.3 & 56.7 \\
 & SVAMP & 67.3 & 90.0 & 68.7 & 77.1 & 89.2 & 78.4 & 71.7 & 80.3 & 73.6 & 70.1 & 93.7 & 70.9 \\
\midrule
Qwen3.5-2B & GSM8K & 73.7 & 92.0 & 74.1 & 74.9 & 85.5 & 77.9 & 78.5 & 89.9 & 79.1 & 74.6 & 93.0 & 75.0 \\
 & SVAMP & 86.7 & 94.7 & 87.3 & 81.8 & 88.6 & 83.1 & 82.0 & 89.3 & 82.3 & 85.0 & 95.0 & 85.3 \\
\bottomrule
\end{tabular}
\end{table*}

\section{Comparison with Stronger Text-Prompt Baselines}
\label{sec:text_prompts}

\begin{table}[!htb]
\centering
\caption{Test accuracy (\%) of text-prompt baselines and ART (three-seed mean). $^{+}$: significantly ahead of ART, $^{\circ}$: no significant difference.}
\label{tab:text_prompts}
\small
\setlength{\tabcolsep}{1.8pt}
\begin{tabular}{llccccc}
\toprule
Model & Input & bAbI & DROP & GSM8K & SVAMP & Tool. \\
\midrule
LFM2-VL-1.6B & Text only & 3.2 & 21.1 & 21.0 & 34.0 & 6.5 \\
 & One-shot & 27.3 & 24.8 & 48.9 & 53.7 & 20.4$^{+}$ \\
 & ART & 40.8 & 40.8 & 60.0 & 73.0 & 8.7 \\
\midrule
Qwen2-VL-2B & Text only & 2.9 & 1.1 & 0.8 & 7.7 & 15.8 \\
 & One-shot & 25.1 & 16.6 & 13.7 & 44.0$^{+}$ & 20.6 \\
 & ART & 45.0 & 39.4 & 40.8 & 37.4 & 53.5 \\
\midrule
Qwen3.5-0.8B & Text only & 23.2 & 33.9 & 55.6 & 67.3 & 51.8 \\
 & One-shot & 32.7 & 22.2 & 43.4 & 59.7 & 58.5 \\
 & GEPA & 49.0$^{+}$ & 45.0 & 49.8 & 65.0 & 83.9$^{+}$ \\
 & ART & 42.1 & 52.6 & 59.4 & 77.1 & 76.2 \\
\midrule
Qwen3.5-2B & Text only & 55.2 & 63.2 & 73.7 & 86.7 & 78.3 \\
 & One-shot & 51.0 & 57.8 & 69.7 & 83.7$^{\circ}$ & 81.0$^{\circ}$ \\
 & ART & 60.4 & 65.6 & 74.9 & 81.8 & 80.5 \\
\bottomrule
\end{tabular}
\end{table}

We compare ART against two text-prompt baselines. 
The text-only baseline already ends with a fixed format instruction (for example \texttt{End your response with: Answer: <number>}). 
The one-shot prompt adds one solved training example before every test question, and we evaluate it on all four models. GEPA \citep{agrawal2026gepa} evolves the instruction itself with Qwen3.5-9B as the reflection model, using $22{,}400$ task-model calls, the same shaped reward, and the same 192 held-out examples that select the ART checkpoint. We ran it on Qwen3.5-0.8B only as Qwen3.5-9B is much more expensive to run compared to the small models. Table~\ref{tab:text_prompts} reports test accuracy, with significance from the paired bootstrap of Appendix~\ref{sec:pairwise}. Every prompt without a marker is significantly behind ART.

\paragraph{Reasoning tasks.} On Qwen3.5-0.8B, ART is significantly ahead of both prompts on GSM8K, SVAMP, and DROP under the paired test, with leads over GEPA of 9.6, 12.1, and 7.6 points. Both prompts are below text only on the math tasks. Across all four models, ART is significantly ahead of the one-shot prompt in 16 of 20 cases and within noise in 2.

\paragraph{ToolMind and bAbI.} GEPA is ahead of ART on ToolMind (83.9\% vs 76.2\%) and bAbI (49.0\% vs 42.1\%), where its prompts grow into rulebooks. These are the two tasks where ART is already behind the best adapter. On bAbI GEPA is also slightly ahead of the best adapter (49.0\% vs 47.7\% for LoRA), while on ToolMind it stays well below the soft prompt (83.9\% vs 98.3\%). The one-shot prompt helps mostly the weak-format models, for example LFM2-VL-1.6B on GSM8K rises from 21.0\% to 48.9\%, and is significantly ahead of ART only on LFM2-VL-1.6B ToolMind and Qwen2-VL-2B-Instruct SVAMP. A GEPA prompt adds up to 6{,}659 characters of context on every request, while ART adds 64--81 visual tokens.

\section{Training}
\label{sec:training_appendix}
\subsection{Training Cost}
\label{sec:training_cost}

Table~\ref{tab:train_cost_full} lists the GPU training time of every training run behind Table~\ref{tab:main_results}. The time counts only the training steps, which include the rollouts and the forward and backward passes. It measures from the end of the step-0 validation to the last logged step, minus the time of every validation in between. Model loading and engine start-up are left out.

\begin{table*}[!t]
\centering
\caption{GPU training time per run in minutes, the sum of the five runs in GPU-hours, mean time per optimization step, and mean completion length in tokens during training.}
\label{tab:train_cost_full}
\footnotesize
\setlength{\tabcolsep}{3pt}
\resizebox{\textwidth}{!}{\begin{tabular}{llcccccccc}
\toprule
\textbf{Model} & \textbf{Method} & \textbf{bAbI} & \textbf{DROP} & \textbf{GSM8K} & \textbf{SVAMP} & \textbf{ToolMind} & \textbf{Total (h)} & \textbf{s/step} & \textbf{Tokens} \\
\midrule
\multirow{3}{*}{LFM2-VL-1.6B} & ART & 23 & 30 & 34 & 28 & 43 & 2.6 & 7.5 & 111 \\
 & Soft prompt & 13 & 21 & 28 & 22 & 23 & 1.8 & 5.1 & 101 \\
 & LoRA & 49 & 77 & 67 & 52 & 135 & 6.3 & 18.3 & 107 \\
\cmidrule{2-10}
\multirow{3}{*}{Qwen2-VL-2B-Instruct} & ART & 25 & 29 & 39 & 28 & 43 & 2.7 & 7.8 & 43 \\
 & Soft prompt & 16 & 18 & 25 & 16 & 29 & 1.7 & 4.9 & 37 \\
 & LoRA & 51 & 74 & 91 & 67 & 150 & 7.2 & 20.8 & 51 \\
\cmidrule{2-10}
\multirow{3}{*}{Qwen3.5-0.8B} & ART & 31 & 40 & 44 & 35 & 37 & 3.1 & 8.8 & 117 \\
 & Soft prompt & 97 & 129 & 87 & 63 & 43 & 7.0 & 19.9 & 196 \\
 & LoRA & 124 & 172 & 165 & 154 & 185 & 13.3 & 38.4 & 134 \\
\cmidrule{2-10}
\multirow{3}{*}{Qwen3.5-2B} & ART & 45 & 50 & 39 & 35 & 51 & 3.7 & 10.4 & 143 \\
 & Soft prompt & 61 & 115 & 90 & 57 & 62 & 6.4 & 18.2 & 205 \\
 & LoRA & 180 & 219 & 181 & 165 & 233 & 16.3 & 47.0 & 150 \\
\midrule
\multirow{2}{*}{Qwen2-VL-7B-Instruct} & ART & 32 & 48 & 59 & 47 & 75 & 4.4 & 12.4 & 62 \\
 & Soft prompt & 34 & 38 & 139 & 37 & 178 & 7.1 & 20.3 & 71 \\
\cmidrule{2-10}
\multirow{2}{*}{GLM-4.6V-Flash} & ART & 82 & 98 & 78 & 46 & 343 & 10.8 & 30.6 & 110 \\
 & Soft prompt & 60 & 75 & 140 & 37 & 246 & 9.3 & 26.4 & 107 \\
\cmidrule{2-10}
\multirow{2}{*}{Gemma-3-12B-it} & ART & 94 & 96 & 84 & 58 & 133 & 7.8 & 22.2 & 98 \\
 & Soft prompt & 173 & 208 & 90 & 65 & 138 & 11.2 & 32.2 & 110 \\
\bottomrule
\end{tabular}}
\end{table*}

\paragraph{Small models.} The soft prompt is the cheapest method on LFM2-VL-1.6B and Qwen2-VL-2B-Instruct. It enters the model after the vision tower, while the ART gradient also passes through it. ART takes 1.5--1.6$\times$ the time of the soft prompt over the five benchmarks. LoRA takes 18--21 seconds per step against 8 seconds for ART, which makes it 2.4--2.6$\times$ slower. The Qwen3.5 runs benefit from being trained with \texttt{flash-linear-attention} and \texttt{causal-conv1d}, which bypasses a slow PyTorch implementation. Over the five benchmarks, the soft prompt takes 1.8--2.2$\times$ and LoRA 4.3--4.4$\times$ the time of ART.

\paragraph{Larger models.} Single runs on the larger models vary more than on the small ones. The step time is steady within a run, so the variation is between runs. On Gemma-3-12B-it the soft prompt is slower than ART on all five benchmarks (65--208 vs 58--133 minutes). On Qwen2-VL-7B-Instruct the soft prompt is about $2\times$ slower than ART on GSM8K and ToolMind (139 vs 59 and 178 vs 75 minutes) but within 2 minutes of ART or faster on the other three benchmarks. The GLM-4.6V-Flash ToolMind ART run took 81 seconds per step, against 11--23 seconds for the other GLM-4.6V-Flash ART runs. A similar slowdown was observed for the soft prompt on the GLM-4.6V-Flash ToolMind run.

\subsection{Accuracy Improvement during Training}
\label{sec:training_dynamics}

\begin{figure*}[!t]
\centering
\includegraphics[width=0.48\textwidth]{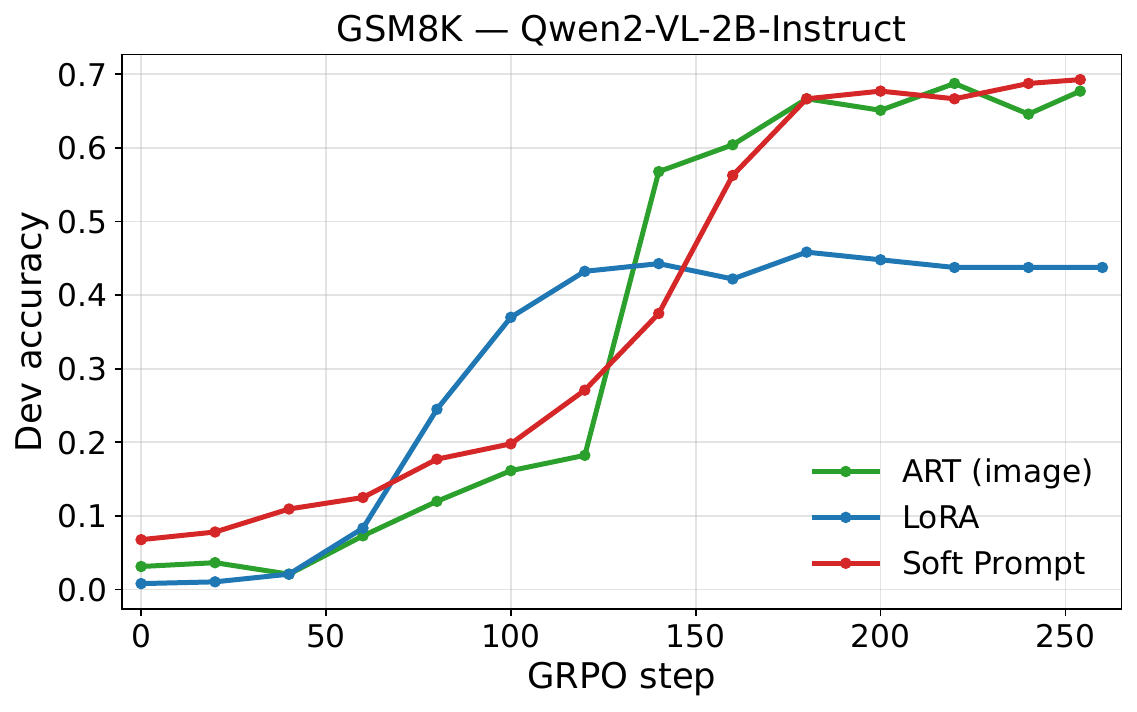}
\includegraphics[width=0.48\textwidth]{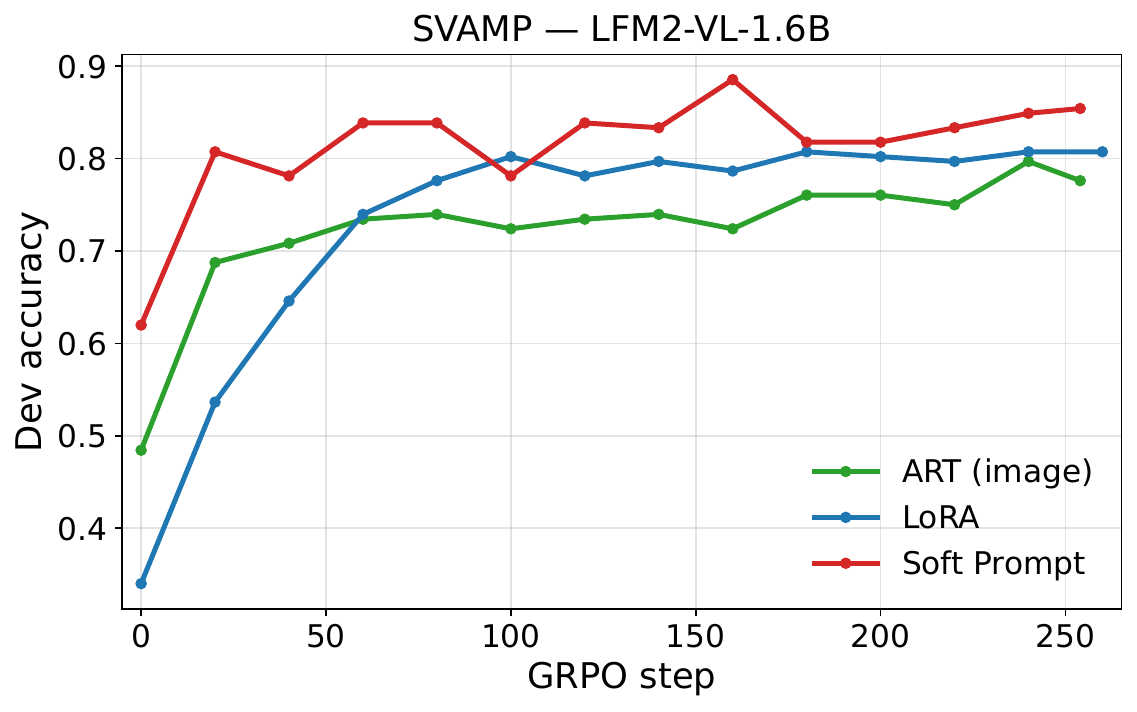}
\caption{Development accuracy over training steps for GSM8K (Qwen2-VL-2B-Instruct, left) and SVAMP (LFM2-VL-1.6B, right).}
\label{fig:devacc_curves}
\end{figure*}

Figure~\ref{fig:devacc_curves} shows accuracy changes during training for the seed-1 runs (strict accuracy on the 192-prompt validation probe, logged every 20 GRPO steps) for GSM8K with Qwen2-VL-2B-Instruct and SVAMP with LFM2-VL-1.6B. On GSM8K, ART and the soft prompt stay low for the first steps and then improve quickly, while on SVAMP accuracy rises steadily.

\begin{figure*}[!t]
\centering
\includegraphics[width=0.48\textwidth]{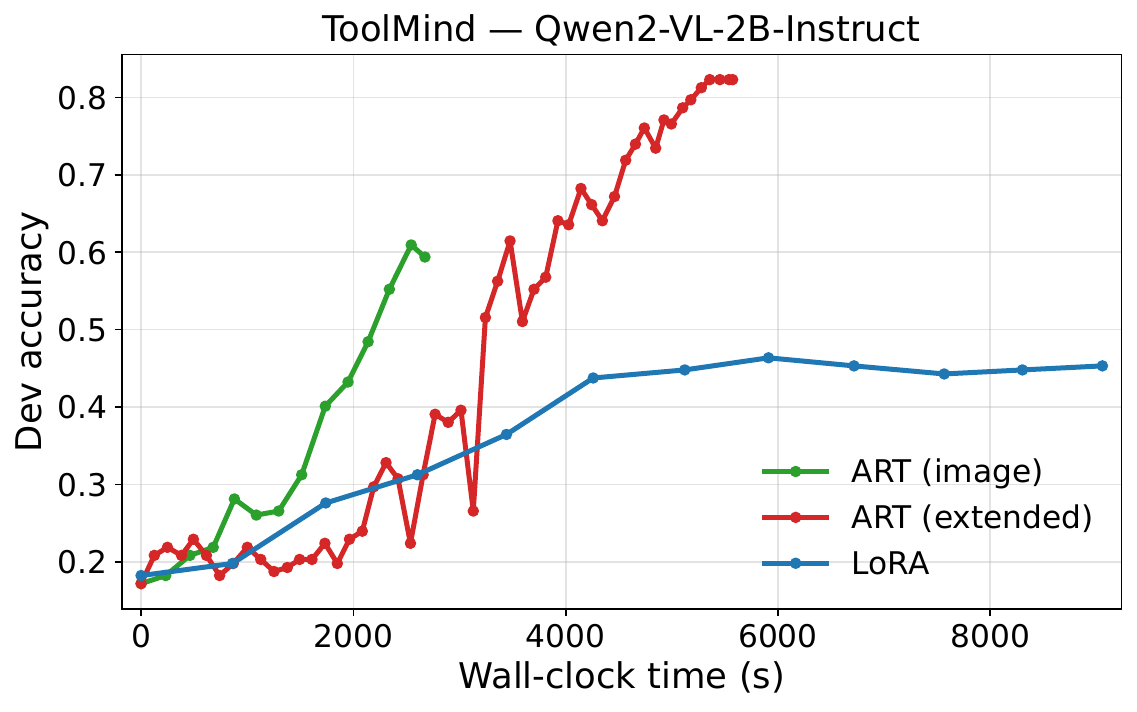}
\includegraphics[width=0.48\textwidth]{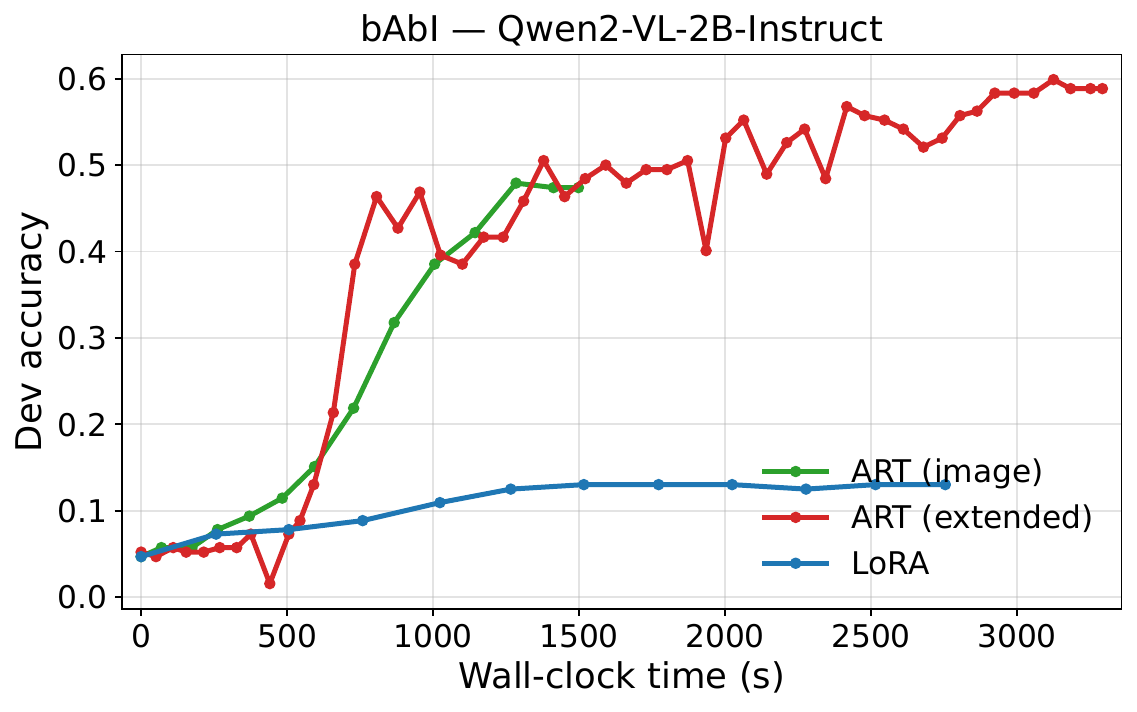}
\caption{Development accuracy over wall-clock time for Qwen2-VL-2B-Instruct on ToolMind (left) and bAbI (right).}
\label{fig:reward_time}
\end{figure*}

Figure~\ref{fig:reward_time} plots validation accuracy against wall-clock time during training on ToolMind and bAbI for Qwen2-VL-2B-Instruct. 
On both benchmarks the ART run finishes before LoRA (about 2{,}700 vs 9{,}100 seconds on ToolMind and 1{,}500 vs 2{,}800 seconds on bAbI) and ends above it ($59\%$ vs $45\%$ and $47\%$ vs $13\%$).
An extended ART run (two epochs with half the effective batch) takes about twice as long as the default one, which is still shorter than LoRA on ToolMind and comparable to it on bAbI.
Note that the test ranking in the main results on ToolMind is reversed, since the 192-example validation probe is not fully representative.

\section{Multi-Tenant Serving Cost}
\label{sec:inference_cost}

We benchmark serving cost under vLLM~0.19.1 on a single NVIDIA A100-40GB and a single V100-32GB. We run Qwen3.5-0.8B and repeat the benchmark on GLM-4.6V-Flash, Qwen2-VL-7B-Instruct, Qwen3.5-9B, and a public 4-bit AutoRound checkpoint of Qwen3.5-27B. All conditions share the model, maximum context, and batch size ($16$). Every request decodes exactly $128$ tokens. We measure over $25$ batches the time-to-first-token (TTFT) and the per-token decode cost (ms/tok) and report their medians. We compute throughput as generated tokens over wall time.

We compare the frozen baseline with ART, a soft prompt of matched token count, and LoRA, each with one artifact shared by the whole batch and with eight \emph{different} artifacts spread across it. The soft prompt is sent through vLLM's \texttt{image\_embeds} input, and the LoRA adapters through vLLM's multi-adapter (S-LoRA-style) path. Tables~\ref{tab:inference_cost}, \ref{tab:inference_cost_large}, and~\ref{tab:inference_cost_qwen2vl} report the results.

\begin{table*}[!t]
\centering
\caption{Serving benchmark on Qwen3.5-0.8B and GLM-4.6V-Flash (vLLM, batch $16$, $128$ decode tokens). $\times$ is throughput relative to text-only (higher is better), TTFT is in ms and ms/tok is the per-token decode cost (lower is better). ``$\times$ 8 tenants'' serves eight different images, soft prompts, or adapters in one batch. Text-only throughput is $4128$ (A100) and $1991$ (V100) tok/s for 0.8B and $949$ and $471$ tok/s for GLM-4.6V-Flash.}
\label{tab:inference_cost}
\footnotesize
\setlength{\tabcolsep}{2.5pt}
\resizebox{\textwidth}{!}{\begin{tabular}{lccccccccccccccc}
\toprule
 & \multicolumn{3}{c}{\textbf{0.8B, A100}} & & \multicolumn{3}{c}{\textbf{0.8B, V100}} & & \multicolumn{3}{c}{\textbf{GLM, A100}} & & \multicolumn{3}{c}{\textbf{GLM, V100}} \\
\cmidrule(lr){2-4}\cmidrule(lr){6-8}\cmidrule(lr){10-12}\cmidrule(lr){14-16}
\textbf{Condition} & \textbf{$\times$} & \textbf{TTFT} & \textbf{ms/tok} & & \textbf{$\times$} & \textbf{TTFT} & \textbf{ms/tok} & & \textbf{$\times$} & \textbf{TTFT} & \textbf{ms/tok} & & \textbf{$\times$} & \textbf{TTFT} & \textbf{ms/tok} \\
\midrule
Baseline (text only)   & 1.00 & 76  & 3.30 & & 1.00 & 252 & 6.08 & & 1.00 & 142 & 15.9 & & 1.00 & 344  & 31.5 \\
ART image              & 0.91 & 122 & 3.33 & & 0.91 & 355 & 6.10 & & 0.92 & 269 & 16.0 & & 0.92 & 682  & 31.8 \\
\quad $\times$ 8 tenants     & 0.91 & 123 & 3.32 & & 0.90 & 357 & 6.24 & & 0.94 & 271 & 16.0 & & 0.92 & 686  & 31.8 \\
Soft prompt            & 0.83 & 153 & 3.36 & & 0.88 & 376 & 6.10 & & 0.93 & 280 & 16.0 & & 0.92 & 682  & 31.9 \\
\quad $\times$ 8 tenants     & 0.83 & 139 & 3.53 & & 0.88 & 367 & 6.30 & & 0.93 & 286 & 16.0 & & 0.92 & 689  & 31.8 \\
LoRA                   & 0.63 & 118 & 5.28 & & 0.26 & 339 & 28.4 & & 0.87 & 166 & 18.1 & & 0.31 & 1332 & 100.6 \\
\quad $\times$ 8 tenants     & 0.49 & 134 & 6.89 & & 0.10 & 363 & 75.7 & & 0.77 & 168 & 20.8 & & 0.07 & 1503 & 465.2 \\
\bottomrule
\end{tabular}}
\end{table*}

\paragraph{Engine settings.} The LoRA engine runs with \texttt{max\_loras}${=}8$, \texttt{max\_lora\_rank}${=}16$, and \texttt{max\_cpu\_loras}${=}16$ and the default LoRA kernels of vLLM~0.19.1. CUDA graphs stay on and prefix caching is off in every condition. Each condition runs four warm-up batches before timing, so adapter loading falls outside the timed window.

\paragraph{Decode cost.} On the 0.8B model, ART and the soft prompt keep $0.83$--$0.91\times$ of the text-only throughput. They add only a one-time prefill cost (a higher TTFT) and leave the per-token decode cost at its baseline value. LoRA instead inflates the \emph{per-token} decode cost, by $60\%$ on the A100 ($5.3$ vs $3.3$~ms/token) and over $4\times$ on the V100 ($28$ vs $6$~ms/token), because the adapter matrices are applied at every decoding step.

\paragraph{Merged adapter.} We merge the GSM8K adapter into the 0.8B weights and serve the result as an ordinary checkpoint without LoRA support, with the same settings on the V100. Its throughput is $1.001\times$ that of the frozen baseline.

\paragraph{Eight tenants.} Serving eight different ART images, or eight different soft prompts, in one batch costs essentially the same as serving one, because each is ordinary input data with no engine-side state. On the 0.8B model, LoRA's multi-adapter path drops to $0.49\times$ (A100) and $0.10\times$ (V100) as the engine manages eight adapters at once.

\paragraph{Larger models.} The larger models use untrained LoRA adapters with the same rank and target modules, untrained soft prompts, and the 0.8B ART images. Serving cost does not depend on the adapter values. The 0.8B pattern holds on every model. ART and the soft prompt keep $0.86$--$0.95\times$ of the text-only throughput, while LoRA raises the per-token decode cost most on the V100, for example from $31.5$ to $100.6$~ms/tok on GLM-4.6V-Flash.

\begin{figure*}[!t]
\centering
\includegraphics[width=0.48\textwidth]{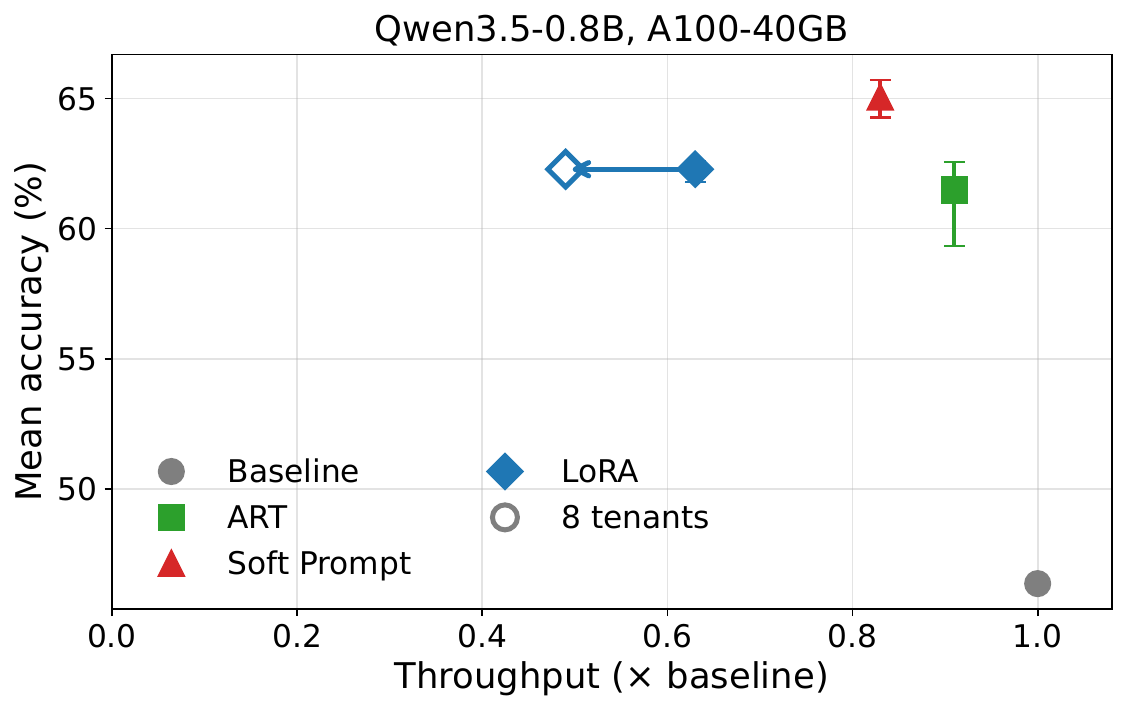}
\includegraphics[width=0.48\textwidth]{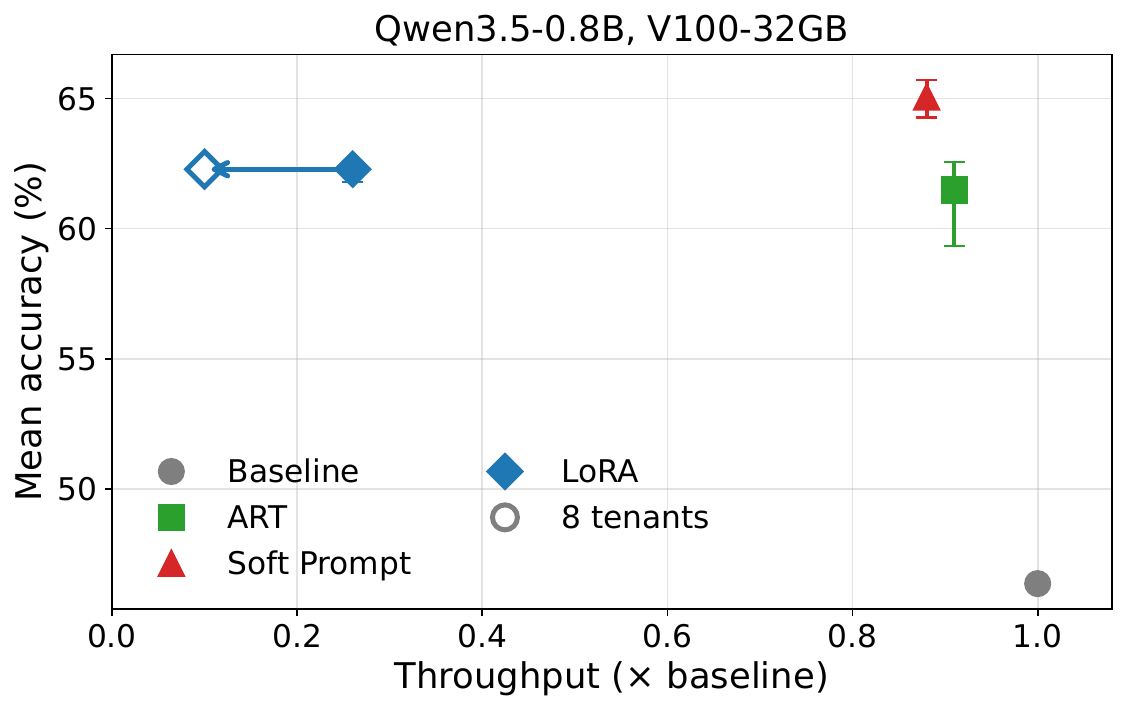}
\caption{Accuracy against relative throughput for Qwen3.5-0.8B on the A100 (left) and V100 (right). Hollow markers: eight tenants.}
\label{fig:acc_tp}
\end{figure*}

\paragraph{Accuracy versus throughput.} Figure~\ref{fig:acc_tp} combines the serving benchmark with the mean Qwen3.5-0.8B accuracy over the five benchmarks and the three training seeds, with bars spanning the lowest and highest seed. The soft prompt ($65.1\%$) is $3.6$ points above ART ($61.5\%$), all of it from ToolMind, at a throughput close to ART's. LoRA ($62.3\%$) is close to ART on accuracy.

\begin{table*}[!t]
\centering
\caption{Serving benchmark on Qwen3.5-9B and 4-bit quantized Qwen3.5-27B, as in Table~\ref{tab:inference_cost}. Text-only throughput is $969$ (A100) and $439$ (V100) tok/s for 9B and $708$ and $155$ tok/s for 27B.}
\label{tab:inference_cost_large}
\footnotesize
\setlength{\tabcolsep}{2.5pt}
\resizebox{\textwidth}{!}{\begin{tabular}{lccccccccccccccc}
\toprule
 & \multicolumn{3}{c}{\textbf{9B, A100}} & & \multicolumn{3}{c}{\textbf{9B, V100}} & & \multicolumn{3}{c}{\textbf{27B, A100}} & & \multicolumn{3}{c}{\textbf{27B, V100}} \\
\cmidrule(lr){2-4}\cmidrule(lr){6-8}\cmidrule(lr){10-12}\cmidrule(lr){14-16}
\textbf{Condition} & \textbf{$\times$} & \textbf{TTFT} & \textbf{ms/tok} & & \textbf{$\times$} & \textbf{TTFT} & \textbf{ms/tok} & & \textbf{$\times$} & \textbf{TTFT} & \textbf{ms/tok} & & \textbf{$\times$} & \textbf{TTFT} & \textbf{ms/tok} \\
\midrule
Baseline (text only)   & 1.00 & 172 & 15.3 & & 1.00 & 672  & 31.3 & & 1.00 & 505 & 18.2 & & 1.00 & 2310 & 85.4 \\
ART image              & 0.95 & 267 & 15.4 & & 0.91 & 1107 & 31.5 & & 0.92 & 896 & 17.7 & & 0.86 & 3836 & 90.3 \\
\quad $+$ 8 tenants     & 0.95 & 280 & 15.4 & & 0.91 & 1104 & 31.5 & & 0.89 & 811 & 18.4 & & 0.86 & 3835 & 90.4 \\
Soft prompt            & 0.95 & 282 & 15.3 & & 0.91 & 1108 & 31.5 & & 0.92 & 783 & 18.5 & & 0.86 & 3840 & 90.3 \\
\quad $+$ 8 tenants     & 0.94 & 281 & 15.4 & & 0.91 & 1109 & 31.5 & & 0.92 & 792 & 18.5 & & 0.86 & 3844 & 90.3 \\
LoRA                   & 0.84 & 187 & 18.3 & & 0.43 & 1106 & 77.1 & & 0.77 & 519 & 24.3 & & 0.48 & 3421 & 189.9 \\
\quad $+$ 8 tenants     & 0.69 & 191 & 22.7 & & 0.15 & 1147 & 237.2 & & 0.57 & 546 & 35.5 & & 0.16 & 3575 & 612.2 \\
\bottomrule
\end{tabular}}
\end{table*}

\begin{table}[!t]
\centering
\caption{Serving benchmark on Qwen2-VL-7B-Instruct, as in Table~\ref{tab:inference_cost}. Text-only throughput is $1194$ (A100) and $561$ (V100) tok/s.}
\label{tab:inference_cost_qwen2vl}
\footnotesize
\setlength{\tabcolsep}{2.5pt}
\begin{tabular}{lccccccc}
\toprule
 & \multicolumn{3}{c}{\textbf{A100}} & & \multicolumn{3}{c}{\textbf{V100}} \\
\cmidrule(lr){2-4}\cmidrule(lr){6-8}
\textbf{Condition} & \textbf{$\times$} & \textbf{TTFT} & \textbf{ms/tok} & & \textbf{$\times$} & \textbf{TTFT} & \textbf{ms/tok} \\
\midrule
Baseline (text only)   & 1.00 & 134 & 12.4 & & 1.00 & 313 & 26.2 \\
ART image              & 0.90 & 239 & 12.6 & & 0.92 & 584 & 26.7 \\
\quad $+$ 8 tenants     & 0.93 & 242 & 12.6 & & 0.92 & 578 & 26.8 \\
Soft prompt            & 0.89 & 347 & 12.4 & & 0.91 & 574 & 26.8 \\
\quad $+$ 8 tenants     & 0.88 & 339 & 12.7 & & 0.91 & 576 & 26.8 \\
LoRA                   & 0.85 & 169 & 14.5 & & 0.46 & 628 & 57.7 \\
\quad $+$ 8 tenants     & 0.67 & 179 & 18.6 & & 0.16 & 650 & 171.8 \\
\bottomrule
\end{tabular}
\end{table}

\section{Cross-Model Transfer}
\label{sec:transfer}

Table~\ref{tab:transfer} reports cross-model transfer within two families, between Qwen3.5-0.8B and Qwen3.5-2B and between Qwen2-VL-2B-Instruct and Qwen2-VL-7B-Instruct. ``2B$\rightarrow$0.8B'' is an image optimized on 2B and evaluated on 0.8B. ``Len.'' is the lenient accuracy introduced in Appendix~\ref{sec:format}.

\begin{table}[!htb]
\centering
\caption{Cross-model transfer of seed-1 ART images on GSM8K and SVAMP (\%).}
\label{tab:transfer}
\small
\setlength{\tabcolsep}{3pt}
\begin{tabular}{llcccc}
\toprule
 & & \multicolumn{2}{c}{\textbf{GSM8K}} & \multicolumn{2}{c}{\textbf{SVAMP}} \\
\cmidrule(lr){3-4}\cmidrule(lr){5-6}
\textbf{Model} & \textbf{Condition} & Strict & Len. & Strict & Len. \\
\midrule
\multirow{3}{*}{Qwen3.5 0.8B}
& Own image & \textbf{60.6} $\pm$ 2.6 & 61.3 & \textbf{76.0} $\pm$ 4.7 & 76.3 \\
& 2B$\rightarrow$0.8B & 38.4 $\pm$ 2.6 & 62.1 & 65.7 $\pm$ 5.5 & 79.0 \\
& Baseline & 55.6 $\pm$ 2.7 & 56.0 & 67.3 $\pm$ 5.3 & 68.7 \\
\midrule
\multirow{3}{*}{Qwen3.5 2B}
& Own image & \textbf{77.0} $\pm$ 2.3 & 78.6 & 82.3 $\pm$ 4.3 & 82.7 \\
& 0.8B$\rightarrow$2B & 0.2 $\pm$ 0.2 & 82.0 & 1.0 $\pm$ 1.2 & 88.0 \\
& Baseline & 73.7 $\pm$ 2.4 & 74.1 & \textbf{86.7} $\pm$ 3.8 & 87.3 \\
\midrule
\multirow{3}{*}{Qwen2-VL 2B}
& Own image & \textbf{47.3} $\pm$ 2.7 & 49.1 & \textbf{37.7} $\pm$ 5.3 & 37.7 \\
& 7B$\rightarrow$2B & 3.0 $\pm$ 0.9 & 26.2 & 12.0 $\pm$ 3.7 & 17.0 \\
& Baseline & 0.8 $\pm$ 0.5 & 39.3 & 7.7 $\pm$ 3.0 & 28.3 \\
\midrule
\multirow{3}{*}{Qwen2-VL 7B}
& Own image & \textbf{78.6} $\pm$ 2.2 & 78.7 & \textbf{85.0} $\pm$ 4.2 & 85.0 \\
& 2B$\rightarrow$7B & 58.2 $\pm$ 2.7 & 76.4 & 37.3 $\pm$ 5.5 & 75.0 \\
& Baseline & 65.3 $\pm$ 2.6 & 76.7 & 42.7 $\pm$ 5.5 & 76.7 \\
\bottomrule
\end{tabular}
\end{table}

Under strict scoring, transfer from 2B to 0.8B falls below the 0.8B baseline on both benchmarks (GSM8K 38.4\% vs 55.6\%, SVAMP 65.7\% vs 67.3\%). Transfer from 0.8B to 2B collapses to near zero (0.2\% and 1.0\%).
Both effects are largely format failures.
With the 0.8B image, the 2B model still ends its output with the correct number on 82.0\% of GSM8K and 88.0\% of SVAMP problems but no longer writes the answer in the required format. In the other direction, the 2B image lifts the 0.8B model's lenient accuracy above its baseline (62.1\% vs 56.0\% on GSM8K).
Within Qwen2-VL, transfer also falls far short of the own image. The 2B image drops 7B GSM8K accuracy below text only ($58.2\%$ vs $65.3\%$). The 7B image gives the 2B model $3.0\%$ on GSM8K and $12.0\%$ on SVAMP, against $47.3\%$ and $37.7\%$ with its own image. Unlike on Qwen3.5, the loss is not only a format failure. Both transferred images stay below even the lenient baseline ($26.2\%$ vs $39.3\%$ on 2B GSM8K and $76.4\%$ vs $76.7\%$ on 7B GSM8K). On DROP, bAbI, and ToolMind both transfers stay within 2.4 points of text only.
This agrees with the poor cross-architecture transfer of adversarial visual prompts \citep{schaeffer2024failures}.

\section{Image Size}
\label{sec:imgsize}

Table~\ref{tab:imgsize} reports ART on GSM8K across image resolutions, with one training run per size and seed 1. The $256$ column repeats the seed-1 main-grid runs.

\begin{table}[!htb]
\centering
\caption{GSM8K accuracy (\%) of ART images across resolutions. Best in \textbf{bold}.}
\label{tab:imgsize}
\small
\setlength{\tabcolsep}{4pt}
\begin{tabular}{lcccccc}
\toprule
\textbf{Model} & \textbf{32} & \textbf{64} & \textbf{128} & \textbf{256} & \textbf{384} & \textbf{512} \\
\midrule
LFM2-VL-1.6B & 53.2 & 57.0 & 58.4 & \textbf{60.0} & 59.6 & 59.1 \\
Qwen2-VL-2B-Instruct & 8.0 & 8.8 & 45.3 & \textbf{47.3} & 46.0 & 39.7 \\
\bottomrule
\end{tabular}
\end{table}

LFM2-VL-1.6B improves with resolution up to $256{\times}256$ and then plateaus.
Qwen2-VL-2B-Instruct is poor at $32{\times}32$ and $64{\times}64$ (8.0--8.8\%), jumps at $128{\times}128$ (45.3\%), stays at 45--47\% up to $384{\times}384$, and dips to 39.7\% at $512{\times}512$.
The low values coincide with the visual token counts in Table~\ref{tab:imgsize_tokens}. Qwen2-VL produces only 5 tokens at $32$ and $64$ and 25 at $128$, whereas LFM2-VL-1.6B and the Qwen3.5 models stay at 64 tokens up to $256$. The LFM2-VL-1.6B gain from $32$ to $256$ therefore comes from pixel detail, not from more tokens.

\begin{table}[!htb]
\centering
\caption{Visual tokens per image size.}
\label{tab:imgsize_tokens}
\small
\setlength{\tabcolsep}{4pt}
\begin{tabular}{lcccccc}
\toprule
\textbf{Model} & \textbf{32} & \textbf{64} & \textbf{128} & \textbf{256} & \textbf{384} & \textbf{512} \\
\midrule
LFM2-VL-1.6B & 64 & 64 & 64 & 64 & 144 & 256 \\
Qwen2-VL-2B-Instruct & 5 & 5 & 25 & 81 & 196 & 324 \\
Qwen3.5-0.8B / 2B & 64 & 64 & 64 & 64 & 144 & 256 \\
\bottomrule
\end{tabular}
\end{table}

\section{Pixel-Value Statistics}
\label{sec:pixel_stats}

\begin{figure*}[!t]
\centering
\includegraphics[width=\textwidth]{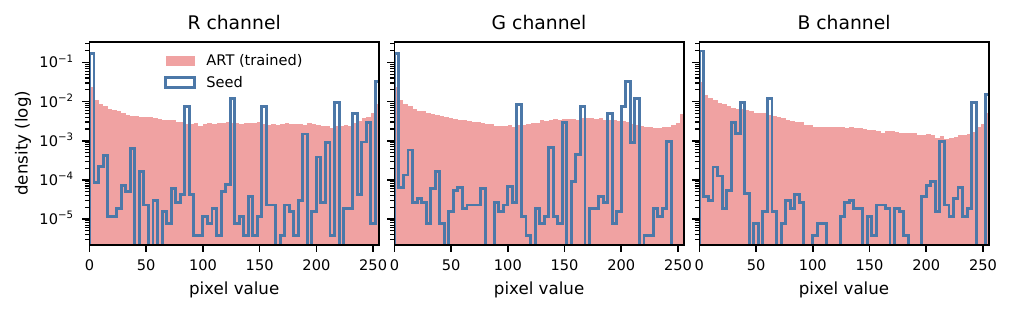}
\caption{Per-channel pixel-value distributions (log density) of the seed image versus the optimized ART image (GSM8K, Qwen3.5-0.8B, seed 1).}
\label{fig:rgb_hist}
\end{figure*}

Figure~\ref{fig:rgb_hist} shows the increase in entropy discussed in Section~\ref{sec:resultsStorage}. The seed image uses only a small number of pixel values, whereas the optimized ART image spreads almost uniformly across the full 8-bit range. Per-channel entropy increases from $\approx 1.3$--$1.9$ bits for the GSM8K seed to $\approx 7.4$--$7.7$ bits (near the 8-bit ceiling) for the optimized image on Qwen3.5-0.8B.
Table~\ref{tab:png_size} lists the PNG size of every seed image and seed-1 ART image at $256{\times}256$. Every ART image takes 165--184~KB, close to the 192~KB of raw pixels, while the seed images take 4--38~KB.

\begin{table}[!htb]
\centering
\caption{PNG file size (KB) of the seed images and the ART images at $256{\times}256$.}
\label{tab:png_size}
\small
\setlength{\tabcolsep}{2pt}
\begin{tabular}{lccccc}
\toprule
\textbf{Image} & \textbf{bAbI} & \textbf{DROP} & \textbf{GSM8K} & \textbf{SVAMP} & \textbf{Tool} \\
\midrule
Seed & 15.5 & 15.5 & 3.9 & 3.9 & 37.7 \\
LFM2-VL-1.6B & 178.3 & 177.2 & 174.7 & 175.0 & 183.6 \\
Qwen2-VL-2B & 170.2 & 171.3 & 165.2 & 167.1 & 179.3 \\
Qwen3.5-0.8B & 179.0 & 179.2 & 173.7 & 173.6 & 181.1 \\
Qwen3.5-2B & 178.8 & 179.8 & 175.2 & 172.1 & 181.7 \\
\bottomrule
\end{tabular}
\end{table}

\section{Distance from the Seed Image}
\label{sec:seed_distance}

ART images move far from their seed (Table~\ref{tab:seed_distance}). Averaged over the 20 seed-1 main-grid images, the largest change of a single pixel value is 94 of 255 levels and the RMS change is 37 levels. Three quarters of all pixel values move by more than 8 levels (the usual $\epsilon{=}8/255$ bound for adversarial images), and 62\% move by more than 16. The PSNR against the seed is 16.8\,dB.

\begin{table}[!htb]
\centering
\caption{Change of ART images relative to their seed in 8-bit levels, averaged over the five benchmarks.}
\label{tab:seed_distance}
\small
\setlength{\tabcolsep}{4pt}
\begin{tabular}{lccccc}
\toprule
Model & $\ell_\infty$ & RMS & PSNR & $>8$ & $>16$ \\
\midrule
Qwen3.5-0.8B & 96 & 36.3 & 17.0 & 74.6\% & 61.3\% \\
Qwen3.5-2B & 89 & 37.1 & 16.8 & 75.3\% & 62.4\% \\
LFM2-VL-1.6B & 97 & 37.0 & 16.8 & 75.2\% & 62.0\% \\
Qwen2-VL-2B-Instr. & 95 & 37.6 & 16.7 & 75.9\% & 63.3\% \\
\bottomrule
\end{tabular}
\end{table}

\end{document}